\documentclass[10pt,twocolumn,letterpaper]{article}

\usepackage[pagenumbers]{cvpr}

\usepackage{graphicx}
\usepackage{mathtools}
\usepackage{amssymb}
\usepackage{booktabs}
\usepackage{multirow}
\usepackage{mathtools}
\usepackage{epsfig}
\usepackage{balance}
\usepackage{hhline}
\usepackage{url}
\usepackage{pifont}
\usepackage{xcolor,soul}
\usepackage{nicefrac}
\usepackage[accsupp]{axessibility}

\usepackage[pagebackref,breaklinks,colorlinks]{hyperref}

\makeatletter
\newcommand*\bigcdot{\mathpalette\bigcdot@{.5}}
\newcommand*\bigcdot@[2]{\mathbin{\vcenter{\hbox{\scalebox{#2}{$\m@th#1\bullet$}}}}}

\newcommand\blfootnote[1]{
  \begingroup
  \renewcommand\thefootnote{}\footnote{#1}
  \addtocounter{footnote}{-1}
  \endgroup
}

\usepackage[capitalize]{cleveref}
\crefname{section}{Sec.}{Secs.}
\Crefname{section}{Section}{Sections}
\Crefname{table}{Table}{Tables}
\crefname{table}{Tab.}{Tabs.}

\begin{document}

\title{Self-Supervised Video Similarity Learning}

\author{Giorgos Kordopatis-Zilos$^{*1}$ \quad Giorgos Tolias$^1$ \quad Christos Tzelepis$^2$ \quad Ioannis Kompatsiaris$^3$ \\ Ioannis Patras$^2$ \quad Symeon Papadopoulos$^3$\\ \\
$^1$VRG, FEE, Czech Technical University in Prague \\
$^2$Queen Mary University of London\\
$^3$Information Technologies Institute, CERTH\\
\vspace{5pt}
{\tt\small \{kordogeo,toliageo\}@fel.cvut.cz \hspace*{0.1em} \{c.tzelepis,i.patras\}@qmul.ac.uk \hspace*{0.1em} \{ikom,papadop\}@iti.gr}
}
\maketitle

\begin{abstract}
We introduce S$^2$VS, a video similarity learning approach with self-supervision. Self-Supervised Learning (SSL) is typically used to train deep models on a proxy task so as to have strong transferability on target tasks after fine-tuning. Here, in contrast to prior work, SSL is used to perform video similarity learning and address multiple retrieval and detection tasks at once with no use of labeled data. This is achieved by learning via instance-discrimination with task-tailored augmentations and the widely used InfoNCE loss together with an additional loss operating jointly on self-similarity and hard-negative similarity. We benchmark our method on tasks where video relevance is defined with varying granularity, ranging from video copies to videos depicting the same incident or event. We learn a single universal model that achieves state-of-the-art performance on all tasks, surpassing previously proposed methods that use labeled data. The code and pretrained models are publicly available at: \url{https://github.com/gkordo/s2vs}
\end{abstract}
\blfootnote{$^*$Research partially conducted at ITI-CERTH.}

\section{Introduction}\label{sec:introduction}
    
    Self-supervised learning is a popular approach, especially for learning representations that are amenable to transfer to different tasks~\cite{chen2020simple,he2020momentum,caron2021emerging,han2020self,qian2021spatiotemporal}. SSL allows to scale-up the dataset size by not relying on manual labeling and is known to obtain representations with high transferability. The commonly studied setup is to consider SSL for pre-training on a proxy task and then perform supervised fine-tuning on different target tasks~\cite{chen2020simple,he2020momentum,caron2021emerging}. In this work, we rather perform SSL and directly use the model on video similarity-related tasks.
    
    \begin{figure}[t]
        \centering
        \includegraphics[width=\linewidth]{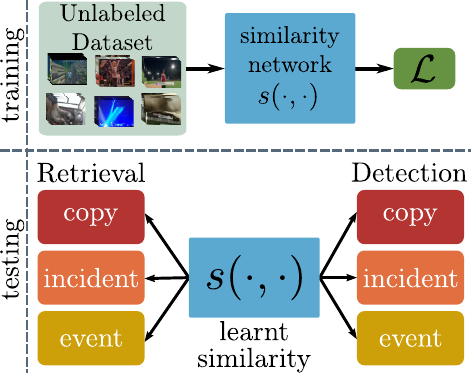}
        \caption{A video similarity network is trained with SSL to compare two videos. The resulting model is used, without any further training, for retrieval and detection of relevant videos in different tasks, where the definition of relevance ranges from video \textit{copies} to videos capturing the same \textit{incidents} and \textit{events}.
        \label{fig:intro}
        \vspace{-5pt}}
    \end{figure}
    
    Computing similarity between videos is a common objective across a number of video retrieval~\cite{wu2007,kordopatis2019a,revaud2013} and video detection~\cite{law2007video,jiang2014} problems. The definition of what is a relevant video to retrieve or detect may differ according to the task at hand. In this work, three cases are considered: i) video copies~\cite{wu2007,jiang2014}, \ie, edited versions of the same source video, ii) videos of the same incident~\cite{kordopatis2019a}, \ie, videos capturing the same spatio-temporal span, and iii) videos of the same event~\cite{revaud2013}, \ie, videos capturing the same spatial or temporal span. In this work, we target both retrieval and detection. In the former, only ranking per query matters; therefore, the distribution of similarities can vary among queries. While in the latter, the ability to apply a similarity threshold and detect relevant videos matters.

    Supervised learning of specialised models per task is very demanding in terms of training data collection, especially in the video domain. Instead, in the proposed method, we are learning a single model via SSL to perform all retrieval and detection tasks (see Figure~\ref{fig:intro}) without further fine-tuning. We inject self-supervision into video similarity learning by adopting the concept of instance-discrimination~\cite{chen2020simple}, where each video forms its own class, and any transformation of it preserves the class label.
    
    In this work, we adopt the ViSiL~\cite{kordopatis2019b} architecture for video similarity, which needs labeled video datasets for its development in prior works~\cite{kordopatis2019b,kordopatis2022}, but we train it in a self-supervised way and argue that instance-discrimination through augmentations is well suited for all the aforementioned tasks. To pronounce the synergy, we develop an appropriate composition of video augmentations and propose a model-tailored loss combined with a standard SSL loss. By eliminating the need for video annotations, we are able to train on large video datasets and achieve state-of-the-art results on all target retrieval and detection tasks. Evaluation is performed on three standard benchmarks, namely, VCDB~\cite{jiang2014}, FIVR~\cite{kordopatis2019a}, and EVVE~\cite{revaud2013}.

    In summary, our contributions include the following:
    \vspace{-7pt}
    \begin{itemize}
    \setlength\itemsep{-4pt} 
    \item We perform SSL via instance-discrimination for video similarity estimation and surpass existing results, obtained with fully supervised training, on three different retrieval and detection tasks.
    \item The performance of the InfoNCE loss~\cite{oord2018representation} is improved by a proposed loss that acts jointly on self-similarity and hard-negative similarity of each video in the batch.
    \item We are the first to jointly benchmark retrieval and detection performance on a range of video-relevance granularities. Additionally, we repurpose the FIVR dataset, whose performance has almost reached saturation, and evaluate only on hard examples.
    \end{itemize}
\section{Related Work}\label{sec:related_work}

    Video similarity and self-supervised learning are the two research fields that are most relevant to our work. 

    \subsection{Video similarity}\label{sec:preliminaries} 
        
        Video similarity methods can be roughly classified into two general categories, \ie, \textit{global representation} and \textit{matching} approaches.
        
        \emph{Global representation} approaches first design or learn a mapping of input examples to a vector space and then use standard distance metrics or similarity measures to compare pairs of examples. These methods reduce down to representation learning, typically called global representation or descriptor, in the sense that the input example is represented by a single vector. Early methods extract hand-crafted features~\cite{huang1999,ojala1996} from all video frames and use aggregation schemes, \eg, mean pooling~\cite{wu2007,huang2010}, Bag-of-Words~\cite{sivic2003,cai2011,shang2010}, to generate global video vectors. More recent approaches use deep features combined with learnable aggregation methods, \ie, using unsupervised schemes~\cite{gao2017,kordopatis2017a,ng2022vrag} or training deep supervised models with metric learning~\cite{kordopatis2017b,lee2018,lee2020}. In addition, several methods extract hash codes for the entire video and measure similarity in the Hamming space~\cite{song2011}. The latter typically train deep networks, such as LSTMs~\cite{hochreiter1997,song2018,yuan2020} or Transformers~\cite{vaswani2017,li2021,li2022dual}, with self-supervised schemes that optimize for the preservation of the video adjacencies from the initial feature space to the Hamming one.
        
        \emph{Matching} approaches represent videos with more than a single vector and involve elaborate similarity estimation schemes, leverage spatio-temporal representations, and exploit video alignment or fine-grained similarity functions. Early methods propose handcrafted solutions to assess similarity through video alignment using Temporal Networks~\cite{tan2009,wang2017}, temporal Hough Voting~\cite{douze2010,jiang2016}, or Dynamic Programming~\cite{chou2015}. Other methods build on the foundations of representation learning to generate spatio-temporal representations with transformer-based networks for temporal aggregation~\cite{shao2021,he2022}, multi-attention networks~\cite{wang2021}, attention-based RNN architectures~\cite{feng2018,bishay2019} or Fourier-based representations~\cite{poullot2015,baraldi2018}. Recent work focuses on video similarity networks that design and learn matching functions to estimate the video-to-video similarity~\cite{kordopatis2019b,kordopatis2022,han2021,jiang2021,he2023transvcl}. The matching function is parametric and learnable in this case. ViSiL~\cite{kordopatis2019b} is among the first methods in this direction; it performs fully-supervised training of a video similarity network to capture fine-grained spatial and temporal structures. Also, Distill-and-Select (DnS)~\cite{kordopatis2022} leverages knowledge distillation to train students using ViSiL as the teacher network. In our work, we adopt the DnS variant of ViSiL and train it with self-supervised learning.

        \begin{figure*}[t]
          \centering
          \includegraphics[width=0.995\linewidth]{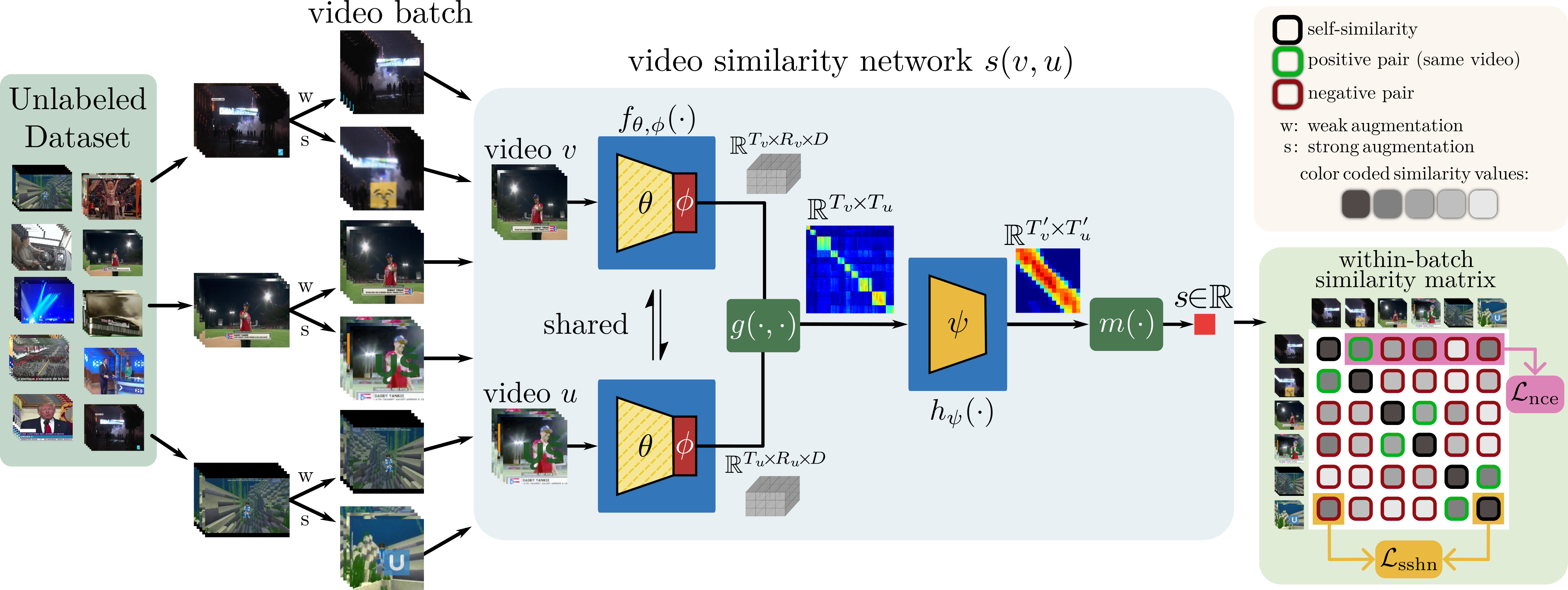}
          \vspace{-5pt}
          \caption{Overview of the proposed approach. Each video in a random batch is augmented twice, and all video pairs are processed by the video similarity network (includes feature representation, spatial matching, and a learnable temporal matching component) to estimate the video-to-video similarity values. Two losses are applied per row of the within-batch similarity matrix: the widely known InfoNCE loss ($\mathcal{L}_\text{nce}$) and the newly proposed $\mathcal{L}_\text{sshn}$ that maximizes self-similarity and minimizes the hardest negative similarity.
          \vspace{-10pt}
          \label{fig:approach_overview}}
        \end{figure*}    
        
    \subsection{Self-supervised learning}\label{sec:self_supervision}
        SSL recently witnessed rapid growth and is leveraged in several vision-related problems. Many examples exist in the image domain for the training of representation models via solving explicit proxy tasks~\cite{doersch2015unsupervised,noroozi2016unsupervised,doersch2017multi,zhang2016colorful,pathak2016context}, discriminating instances through contrastive learning~\cite{chen2020simple,he2020momentum,henaff2020data,tian2020contrastive}, optimizing clustering and representation ~\cite{caron2018deep,caron2020unsupervised,asano2019self}, bootstrapping knowledge with self-distillation~\cite{chen2021exploring,grill2020bootstrap,caron2021emerging} or image reconstruction with masked autoencoders~\cite{bao2021beit,he2022masked}.

        The video domain offers additional avenues for self-supervision, \eg, by exploiting spatio-temporal information, such as frame ordering~\cite{misra2016shuffle,fernando2017self}, motion~\cite{agrawal2015learning,jayaraman2015learning}, multi-modal co-training~\cite{han2020self}, temporal field of view~\cite{recasens2021broaden} or, more recently, video masking autoencoders~\cite{wang2022bevt,tong2022videomae}.The roadmap of augmentations designed for videos has been adopted by some approaches~\cite{qian2021spatiotemporal,kim2019self,han2020self} that train a video representations network on a proxy task. Qian \etal~\cite{qian2021spatiotemporal} use temporally consistent spatial augmentation and contrastive learning. These methods learn a mapping of videos to a vector space, and, most of the time, the goal is to perform fine-tuning on other tasks with good generalization.

        In another line of research, when video-to-video similarity estimation is the objective, mapping to a vector space is not the most suitable choice; instead, a matching function for a video pair is typically the preferred choice. In contrast to this work, where we learn a parametric matching function, prior work uses hand-crafted matching and only learns the representation~\cite{baraldi2018,he2022}. This is the case for near-duplicate video retrieval~\cite{he2022} and video matching through alignment~\cite{baraldi2018}, where self-supervision comes in the form of pre-generated static training datasets through spatio-temporal augmentations. In the same way, He~\etal~\cite{han2021} target video copy localization and, through self-supervision, generate ground truth masks at the level of frame-to-frame correspondences. In contrast to them, we optimize a more general video similarity model and effectively employ it to tackle multiple retrieval and detection tasks. \looseness=-1
        
        Lastly, a related work~\cite{pizzi2022} in the image domain proposes a self-supervised method reflecting the objectives of the target task based on task-specific augmentations. Their method relies on contrastive image representation learning using advanced augmentations, \eg, text and emoji overlays, strong blurring, and CutMix~\cite{yun2019}, and an adopted InfoNCE loss~\cite{oord2018representation}. We use similar task-specific augmentations and losses in our work for videos, instead of images. \looseness=-1

\section{SSL for Video Similarity}\label{sec:proposed_method}
    Our aim is to learn a video similarity function $s\colon \mathcal{V} \times \mathcal{V} \to \mathbb{R}$, where $\mathcal{V}$ is the space of all videos. The goal is for two videos to have high similarity if they are relevant, and low otherwise. The definition of relevance is task-dependent. In our experiments, we consider several evaluation tasks, where relevance ranges from video copies to videos of the same physical event. Nevertheless, we perform training in a single universal way without video labels for supervision. We perform training with self-supervision in the spirit of instance-discrimination, \ie, two augmented videos originating from the same original video are considered as positive to each other, or negative otherwise. In some parts, we follow the work of Pizzi \etal~\cite{pizzi2022}, who perform SSL for image copy detection. The overview of the proposed approach is illustrated in Figure~\ref{fig:approach_overview}. 

    \begin{figure*}[t]
        \centering
                    \centering
            \setlength\tabcolsep{1pt}
            \begin{tabular}{c c c c c c c c}
              & \hspace{-2pt}
              \begin{tabular}{p{2.5cm}}
                   \centering 
                   \textbf{identity} \\ 
                   \textbf{(self-similarity)}
              \end{tabular}
              & \hspace{-2pt}
              \begin{tabular}{p{2.5cm}}
                   \centering 
                   \textbf{weak} \\
                   \textbf{transformations}
              \end{tabular} 
              & \hspace{-2pt}
              \begin{tabular}{p{2.5cm}}
                   \centering 
                   \textbf{global} \\
                   \textbf{transformations}
              \end{tabular}
              & \hspace{-2pt}
              \begin{tabular}{p{2.5cm}}
                   \centering 
                   \textbf{frame} \\
                   \textbf{transformations}
              \end{tabular}
              & \hspace{-2pt}
              \begin{tabular}{p{2.5cm}}
                   \centering 
                   \textbf{temporal} \\
                   \textbf{transformations}
              \end{tabular} &
              \textbf{video-in-video} \\
              & 
              \includegraphics[scale=.11]{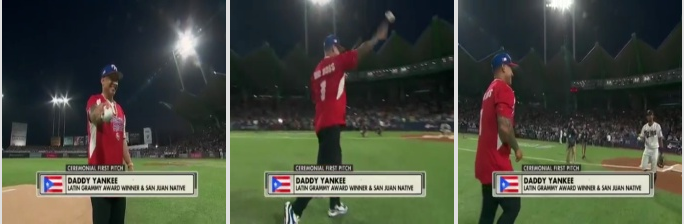} &
              \includegraphics[scale=.11]{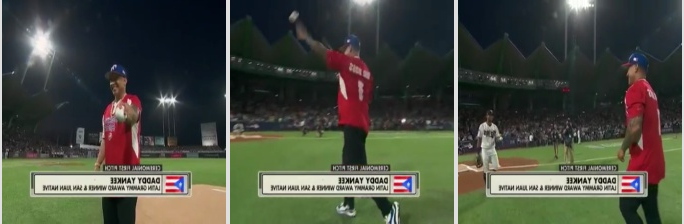} &
              \includegraphics[scale=.11]{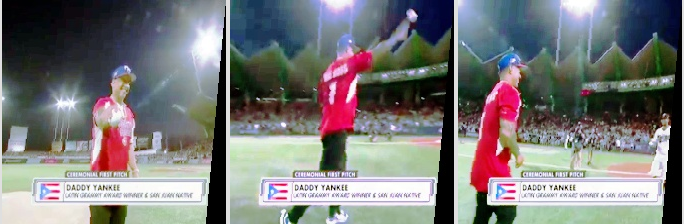} &
              \includegraphics[scale=.11]{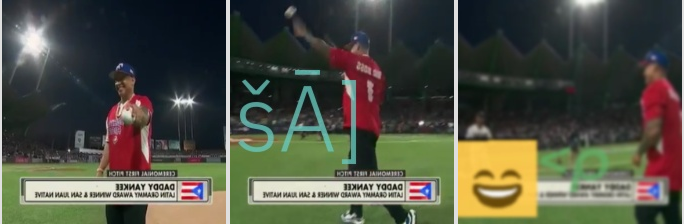} &
              \includegraphics[scale=.11]{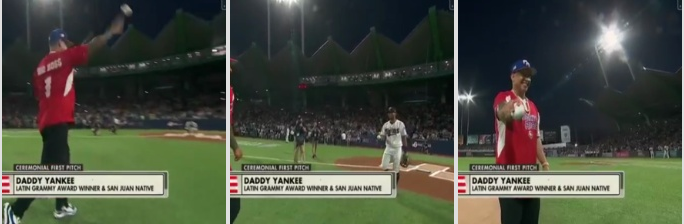} &
              \includegraphics[scale=.11]{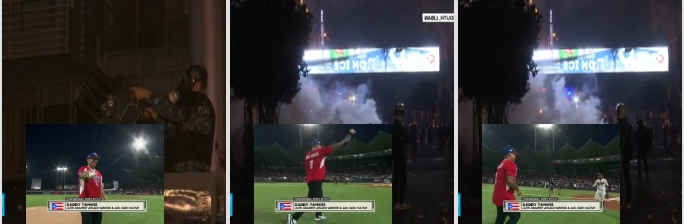}
              \\
              \includegraphics[scale=.11]{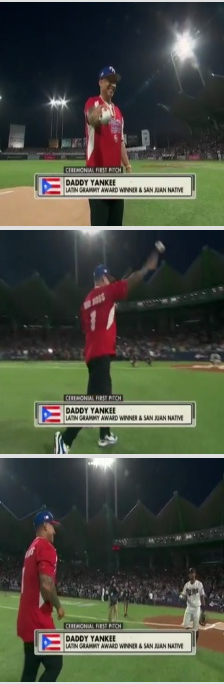} & 
              \includegraphics[scale=.35]{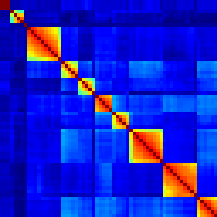} &
              \includegraphics[scale=.35]{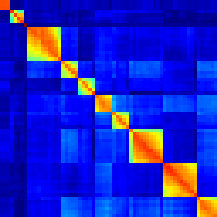} &
              \includegraphics[scale=.35]{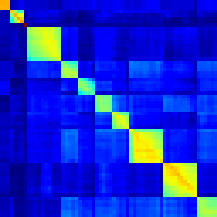} &
              \includegraphics[scale=.35]{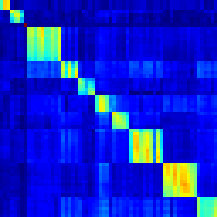} &
              \includegraphics[scale=.35]{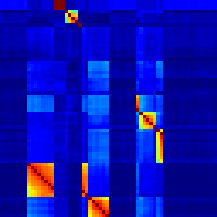} &
              \includegraphics[scale=.35]{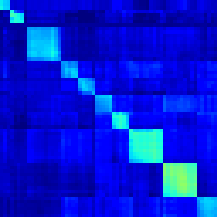}
              \\
              &
              \includegraphics[scale=.23]{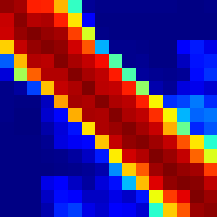} &
              \includegraphics[scale=.23]{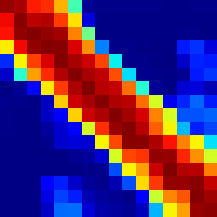} &
              \includegraphics[scale=.23]{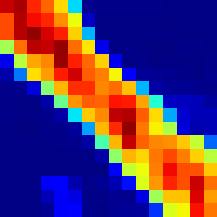} &
              \includegraphics[scale=.23]{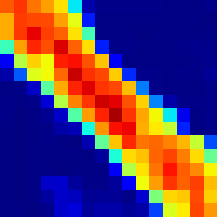} &
              \includegraphics[scale=.23]{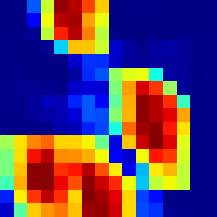} &
              \includegraphics[scale=.23]{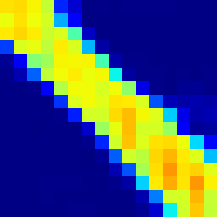}
              \\
            \end{tabular}
        \vspace{-5pt}
        \caption{Temporal similarity matrices formed between the original video and its strongly augmented counterpart. Sampled frames (top), similarity matrices that are the input to (middle), and output of (bottom) the temporal matching function of the proposed S$^2$VS are shown. Only one type of augmentation is applied in each case. All matrices are scaled in [0,1] with blue (red), indicating similarity close to 0 (1).
        \vspace{-12pt}
        \label{fig:sim_mat_augs}}
    \end{figure*}

\subsection{Similarity network}\label{sec:model}
    We adopt the ViSiL variant proposed in DnS~\cite{kordopatis2022}, namely the fine-grained attention student, as our similarity network architecture. It consists of a representation network, a hand-crafted spatial matching function, a learnable temporal matching function, and a final hand-crafted matching function that estimates the final video-level similarity.
    
    The representation network $f_{\theta,\phi}\colon \mathcal{V} \to \mathbb{R}^{T\times R\times D}$ maps an input video to a $D$-dimensional vector per region, for $R$ regions per frame, for $T$ frames, where $R$ and $T$ vary according to the frames' size and video length, respectively. This network consists of a pre-trained backbone network and has a parameter set $\theta$ that is fixed in this work, similar to the prior ones~\cite{baraldi2018,kordopatis2019a,jiang2021}. The learnable part corresponds to the parameter set $\phi$, a dot-attention scheme~\cite{yang2016} that is applied to weigh region vectors based on their saliency.
    
    Given two input videos $v$ and $u$ and their corresponding representations, the hand-crafted spatial matching is performed by the function $g\colon \mathbb{R}^{T_v \times R_v \times D} \times \mathbb{R}^{T_u \times R_u \times D} \to \mathbb{R}^{T_v \times T_u}$, that takes as input two video representations and estimates the temporal similarity matrix. It computes the $R_v \times R_u$ spatial similarity matrix for all frame pairs and then applies Chamfer similarity on each of them to estimate the frame-to-frame similarity. \looseness=-1
    
    The temporal matching is performed by function $h_\psi: \mathbb{R}^{T_v \times T_u} \rightarrow \mathbb{R}^{T^\prime_v \times T^\prime_u}$. This is a four-layer CNN that learns to capture temporal patterns in the input similarity matrices. It outputs a filtered temporal similarity matrix. It holds that $T_v=4T^\prime_v$, and similarly for $u$, due to the CNN design that contains strided max pooling operations. The parameters of the CNN, denoted by $\psi$, are learnable.
    
    Chamfer similarity is applied and denoted by the function $m\colon \mathbb{R}^{T^\prime_v \times T^\prime_u} \to \mathbb{R}$, taking as input the filtered temporal similarity matrix and estimating the final video-level similarity, \ie, the scalar similarity between the two videos.
    
    To summarize, similarity $s(v,u)$, for the video pair consisting of videos $v$ and $u$, is equivalent to $s(v,u)=m\left(h_\psi \left(g(f_{\theta,\phi}(v),f_{\theta,\phi}(u))\right)\right)$, and the goal in this work is to learn $\phi$ and $\psi$ with self-supervision on videos, while $\theta$ remains fixed and is obtained from supervised pre-training on ImageNet. The reader is referred to the original ViSiL work~\cite{kordopatis2019b} for additional details. \vspace{0.5pt}

\subsection{Weak/strong video augmentations}\label{sec:augmentations} \vspace{0.5pt}
    We apply two sets of augmentations to generate two corresponding versions of a training video, \ie, one weakly and one strongly augmented version. Formally, given an original video $v$, the output of an augmentation function $A$ is a video tensor $\tilde{v}=A(v)\in\mathbb{R}^{T_B\times H_B\times W_B\times 3}$, where $T_B$, $H_B$, and $W_B$ correspond to the number of frames, height, and width of the video in the batch, respectively.
    
    \textbf{Weak augmentations} consist of conventional geometric transformations (i.e., resized crop and horizontal flip), applied globally on the entire video, and temporal cropping to select $T_B$ consecutive frames.
      
    \textbf{Strong augmentations} consist of the weak augmentations and several other transformations grouped into the following four categories:
    
    \emph{Global transformations} are frame transformations applied to all frames in a consistent way. We use RandAugment~\cite{cubuk2020}, an automatic augmentation strategy that includes different geometric and photometric image transformations and requires two hyperparameters, namely $N_{RAug}$ and $M_{RAug}$. These correspond to the number of randomly-applied consecutive transformations and their magnitude value that determines their severity, respectively.
      
    \emph{Frame transformations} are applied independently per frame. We use overlay and blurring transformation\footnote{The RandAugment implementation we use does not contain blurring operations. Hence, global transformations do not blur videos.}. Following advanced augmentations from prior work~\cite{pizzi2022}, we add random emojis and text, each with probability $p_{overlay}$, and blur frames with probability $p_{blur}$. We opt for these operations to emulate common video copy transformations.
      
    \emph{Temporal transformations} act only on the temporal dimension and include five operations, with one applied per video. Following \cite{kordopatis2019b}, we use fast forward, slow motion, reverse play, and frame pause, where a single frame is duplicated several times consecutively. In addition, we propose Temporal Shuffle-Dropout (TSD) to alter the global temporal structure but preserve the local one. The video is first split into short clips, each of them with length randomly chosen in $[4,\ldots,T_B/2]$. In the shuffling phase, applied with probability $p_{shuf}$, the clip order is shuffled. In the dropout phase, a clip is dropped with probability $p_{drop}$, where it is either discarded or filled with empty frames or Gaussian noise with probability $p_{cont}$.

    \emph{Video-in-video} randomly mixes two strongly augmented videos, the \textit{host} and the \textit{donor}, in the same batch. The donor video is randomly spatially down-sampled with a factor $\lambda_{viv}$ and is overlaid in a random location within the host video. Each strongly augmented video is chosen as donor with probability $p_{viv}$. Then, a host video is randomly chosen, while the mixed output replaces the donor video. This process requires properly adjusting the instance-discrimination labels since the generated video is the outcome of two others. Video-in-video transformation is very common in real-life video cases.
    
    Figure~\ref{fig:sim_mat_augs} presents the impact of each type of augmentation on the temporal similarity matrices. \emph{Self-similarity}, \ie, identity augmentation, is shown as a reference and hints about the temporal structure of the video. \emph{Weak} augmentation only slightly affects, while \emph{global} and \emph{frame} augmentations noticeably affect the strength on the block diagonal structures. Such a structure is preserved with global transformations but not so much with the frame ones; observe some blue vertical lines indicating a significant impact on the frame representation. Nevertheless, the trained network handles both cases robustly, assigning large similarity values on the diagonal part, as seen on the filtered matrices. The \emph{temporal} transformations significantly alter the global structure but partly preserve the local one, while the video-in-video transformation has a substantial impact on the intensity of the main diagonal, highlighting its challenging aspect; yet, the trained network effectively learns to handle such cases. \looseness=-1

\subsection{Loss on video similarity}\label{sec:losses}
    A random set of $N$ videos, where each video is augmented once with the weak and once with the strong augmentations, forms a training batch of size $B=2N$ denoted by $\mathcal{B} = [v_1, \cdots, v_{2N}]$. We compute the similarity matrix $S \in [0,1]^{B \times B}$, with elements $S_{i,j}=s(v_i,v_j)$, comprising all pairwise video similarities within the batch. Each row of $S$ consists of the self-similarity on the diagonal, one positive-pair similarity, and $B-2$ negative-pair similarities\footnote{This is the case where video-in-video augmentation is not used; otherwise, there can be more (less) positives (negatives).}. Note that $S$ is not symmetric and that the diagonal elements are not equal to 1 because of $h_\psi$. For the $i$-th row of the similarity matrix, let $p(i)$ be the set of column indices of the positive pairs. Additionally, for the $i$-th row, let $n(i)$ be the set of column indices of the negative pairs.
    
    The total loss is a combination of two losses that optimize different parts of $S$: (i) the widely used InfoNCE~\cite{oord2018representation} loss estimated per row excluding the self-similarity value, and (ii) a loss that maximizes the self-similarity, \ie, main diagonal, and minimizes the similarity with the hardest negative, \ie, the negative with the highest similarity, for each video in the batch. \looseness=-1
    
    \textbf{InfoNCE loss} is estimated for each positive pair by
    \begin{equation}
      \mathcal{L}_{\text{nce}}(i,j) = -\log \frac{\exp (S_{i,j} / \tau)}{\exp (S_{i,j} / \tau)+\sum_{k\notin p(i) \cup i } {\exp(S_{i,k} / \tau)}},
    \end{equation}
    where $\tau$ is a temperature hyper-parameter and $(i,j)$ is a positive pair. The final InfoNCE loss is given by the average over all positive pairs as
    \begin{equation}   
    \mathcal{L}_{\text{nce}} = \nicefrac{1}{P} \sum_i \sum_{j\in p(i)} \mathcal{L}_{\text{nce}}(i,j),
    \end{equation} 
    where $P$ is the total number of positive pairs in the batch.
    
    \textbf{Self-similarity -- hardest negative loss:} 
    Since the self-similarity is not equal to 1 by design, we add a loss term that is trying to push it to high values. 
    Together with that, an additional term pushes the hardest negative of each row to have small similarity. 
    For the $i$-th row, this loss is given by \looseness=-1
    \begin{equation}
      \mathcal{L}_{\text{sshn}}(i) = - \underbrace{\log \left(S_{i,i}\right)}_{self-sim} - \underbrace{\log \max_{j \notin p(i)\cup i} \left(1- S_{i,j}\right)}_{hard-negative\ sim},  
    \end{equation}
    and the total loss is given by the average over rows as $\mathcal{L}_{\text{sshn}} = \nicefrac{1}{B} \sum_i \mathcal{L}_{\text{sshn}}(i)$. Note that the hard-negative term resembles entropy maximization through the Kozachenko-Leononenko estimator and a consequent spreading of elements in the representation space~\cite{sablayrolles2018}. Differently to them, we perform this directly on pairwise similarities and not on distances over a vector space.
    
    To this end, we optimize a weighted sum of the losses presented above, as follows
    \begin{equation}
      \mathcal{L} = \mathcal{L}_{\text{nce}} + \lambda\mathcal{L}_{\text{sshn}},
    \end{equation}
    where $\lambda$ is a hyperparameter that tunes the impact of $\mathcal{L}_{\text{sshn}}$.

\section{Evaluation setup}\label{sec:evaluation_setup}
    Here, we present the training/evaluation datasets, the evaluation metrics, and some implementation details.
    
    \subsection{Datasets}\label{sec:datasets}

        \textbf{DnS-100K}~\cite{kordopatis2022} consists of 115,792 unlabeled videos. It is used for knowledge distillation in the original work, but we use it as a training set.
		
		\textbf{VCSL}~\cite{he2022vcsl} is originally created for video copy localization. It contains 9,207 videos with more than 281K copied segments split into training, validation, and test set. Due to the unavailability of several videos, we managed to collect only 8,384 videos. We use this dataset to train our model in a supervised way, only to provide an indicative comparison with the proposed SSL approach.

        \textbf{VCDB} \cite{jiang2014} is created for partial video copy detection. The core dataset ($\mathcal{C}$) contains 528 videos from 28 discrete sets with over 9,000 copied segments. It also contains a set $\mathcal{D}$ of 100,000 distractor videos. We use this dataset for evaluation for detection and retrieval of video copies, considering as related the videos that share at least one copied segment. Moreover, we use the distractor set as an alternative unlabeled training set. We use VCDB, VCDB ($\mathcal{D}$), or VCDB ($\mathcal{C}$+$\mathcal{D}$) to indicate that only set $\mathcal{C}$, only set $\mathcal{D}$, or both sets are used, respectively.
    
        \textbf{FIVR-200K} \cite{kordopatis2019a} is used as a benchmark for fine-grained incident video retrieval. It consists of 225,960 videos and 100 queries. FIVR-200K includes three different subtasks: a) Duplicate Scene Video Retrieval (DSVR), b) Complementary Scene Video Retrieval (CSVR), and c) Incident Scene Video Retrieval (ISVR). In this work, we use the same subsets to evaluate for the corresponding detection tasks, denoted by DSVD, CSVD, and ISVD. For quick comparisons, we also use \textbf{FIVR-5K}~\cite{kordopatis2019b}, a subset of FIVR-200K. We use it in our ablations, denoted by \textbf{FIVR}, where the average performance of the three subtasks is reported.
                
        \textbf{EVVE} \cite{revaud2013} is a dataset for video retrieval. It consists of 620 queries and 2,375 database videos. Due to the unavailability of several videos, we use only 504 queries and 1906 database videos~\cite{kordopatis2019b}, which is roughly $\approx$80\% of the initial dataset. All reported methods are evaluated on this subset.
        
        In summary, we train on DnS-100K, or VCDB($\mathcal{D}$), and evaluate on VCDB for video copies, on FIVR for video copies, and incidents, and on EVVE for video copies, incidents, and events.
        
    \subsection{Evaluation metrics}\label{sec:eval_metrics}
    	To evaluate methods for retrieval, we use mean Average Precision (mAP). AP is equivalent to the area under the precision-recall curve for a particular query, and mAP is obtained by simply averaging over all queries. To evaluate for detection, we use micro Average Precision ($\mu$AP) as a good indicator of detection performance also used in prior work~\cite{law2007video,pizzi2022,douze2021}. This is equivalent to the area under the precision-recall curve for all queries jointly. The lists of similarities from all queries are merged, and the labels defining relevant/non-relevant videos (despite being defined with respect to different queries) are used to estimate precision and recall. All the metrics are re-scaled to the $[0,100]$ range. For retrieval and mAP, only the ranking per query matters; therefore, the distribution of similarities can vary a lot among queries. This is not the case for detection and $\mu$AP, where the ranking among all queries jointly matters, reflecting the ability to apply a threshold and detect the relevant items. \looseness=-1
    
    \subsection{Implementation details}\label{sec:implementation_details}
        \textbf{Pre-trained backbone networks ($\theta$)}: To implement our representation network $f_{\theta,\phi}$, we follow the literature~\cite{kordopatis2019b,shao2021,kordopatis2022} and employ a ResNet50~\cite{he2016} network pretrained on ImageNet~\cite{deng2009imagenet}. We also extract region vectors applying regional max activation pooling~\cite{tolias2016} on intermediate layers~\cite{kordopatis2017a}, whitened through a PCA-whitening layer~\cite{chum2007} learned from 1M region vectors sampled from the VCDB~\cite{jiang2014} dataset.
        
        \textbf{Training process}: We train our network for 30K iterations with a batch size of 64. We employ AdamW~\cite{loshchilov2018decoupled} optimization with learning rate $5\cdot10^{-5}$ and weight decay 0.01. We use cosine learning rate decay with 1K iterations warm-up~\cite{loshchilov2016sgdr}. Other parameters are set to $T_B=32$, $\tau=0.03$, and $\lambda=3$. Also, following the original ViSiL work~\cite{kordopatis2019b}, the similarity regularization is employed with a factor $r=1$. The similarity network generates scores in $[-1,1]$, and we rescale them to $[0,1]$ for the loss calculation. For further implementation details and the complete list of hyperparameter values, we point readers to the supplementary material.

\begin{table*}[t]
    \centering
    \def\arraystretch{1.1}
\setlength\tabcolsep{3.5pt}
\begin{tabular}{l cc ccccc ccccc}
    &  &  & \multicolumn{5}{c}{\textbf{Retrieval}}  & \multicolumn{5}{c}{\textbf{Detection}}  \\  \cmidrule(lr){4-8} \cmidrule(lr){9-13}
    &  &  & \multirow{2}{*}{
    \begin{tabular}{p{.8cm}}
        \centering 
        \vspace{-5pt}
        \textbf{VCDB} \\
        ($\mathcal{C}$+$\mathcal{D}$)
    \end{tabular}}  & 
    \multicolumn{3}{c}{\textbf{FIVR-200K}} & \multirow{2}{*}{\textbf{EVVE}} & \multirow{2}{*}{
    \begin{tabular}{p{.8cm}}
        \centering 
        \vspace{-5pt}
        \textbf{VCDB} \\
        ($\mathcal{C}$+$\mathcal{D}$)
    \end{tabular}} &
    \multicolumn{3}{c}{\textbf{FIVR-200K}} &  \multirow{2}{*}{\textbf{EVVE}} \\ \cmidrule(lr){5-7} \cmidrule(lr){10-12}
    \textbf{Approach} & \textbf{Lab.} & \textbf{Trainset} &  & DSVR & CSVR & ISVR &  & & DSVD & CSVD & ISVD & \\ \midrule
    \textbf{DML}~\cite{kordopatis2017b}         & \checkmark & VCDB ($\mathcal{C}$+$\mathcal{D}$)      & -    & 52.8 & 51.4 & 44.0 & 61.1 & -    & 39.0 & 36.5 & 30.0 & 75.5   \\ 
    \textbf{LAMV}~\cite{baraldi2018}            & \ding{55}  & YFCC100M  & 78.6 & 61.9 & 58.7 & 47.9 & 62.0 & 62.0 & 55.4 & 50.0 & 38.8 & \underline{80.6}    \\ 
    \textbf{TCA$_f$}~\cite{shao2021}            & \checkmark & VCDB ($\mathcal{C}$+$\mathcal{D}$)      & -    & 87.7 & 83.0 & 70.3 & -    & -    & -    & -    & -    & -      \\
    \textbf{VRL$_f$}~\cite{he2022}              & \ding{55}  & internal  & -    & 90.0 & 85.8 & 70.9 & -    & -    & -    & -    & -    & -      \\
    \textbf{ViSiL$_f$}~\cite{kordopatis2019b}   & \ding{55}  & -         & 82.0 & 89.0 & 84.8 & 72.1 & 62.7 & 40.9 & 66.9 & 59.5 & 45.9 & 74.6  \\
    \textbf{ViSiL$_v$}~\cite{kordopatis2019b}   & \checkmark & VCDB ($\mathcal{C}$+$\mathcal{D}$)      & -    & 89.9 & 85.4 & 72.3 & 65.8 & -    & 75.8 & 69.0 & 53.0 & 79.1   \\ 
    \textbf{DnS}~\cite{kordopatis2022}          & \checkmark & DnS-100K  & \textbf{87.9} & 92.1 & 87.5 & \underline{74.1} & 65.1 & \textbf{74.0} & 79.7 & 69.5 & 54.2 & 74.3   \\ \midrule
    \textbf{S$^2$VS} (Ours)                     & \ding{55}  & VCDB ($\mathcal{D}$)  & -   & \textbf{92.7} & \textbf{87.9} & \textbf{74.6} & \textbf{67.2} &  - & \underline{85.7} & \underline{76.9} & \underline{62.8} & \textbf{80.7}   \\
    \textbf{S$^2$VS} (Ours)                     & \ding{55}  & DnS-100K  & \textbf{87.9} & \underline{92.5} & \underline{87.8} & 73.9 & \underline{65.9} & \underline{73.0} & \textbf{89.3} & \textbf{80.2} & \textbf{64.9} & 78.9   \\
\end{tabular}
    \vspace{-5pt}
    \caption{State-of-the-art comparison via retrieval mAP (\%) and detection $\mu$AP (\%) on three evaluation datasets. \textbf{Bold} and \underline{underline} indicate the best and second best approach, respectively. Missing values are either due to unavailability or unfair comparison due to leak of evaluation data during training.
    \vspace{-10pt}}
    \label{tab:mAP_comparison}
\end{table*}

\section{Experiments}\label{sec:experiments}
    We evaluate the performance of the proposed approach on different retrieval and detection tasks related to video similarity, compare its performance to the state-of-the-art methods, and conduct an ablation study.

    \subsection{Comparison with the state-of-the-art}\label{sec:sota}
         
        We compare the proposed S$^2$VS method with the following approaches. \textbf{DML}~\cite{kordopatis2017b} extracts a video embedding based on a network trained with supervised deep metric learning. \textbf{LAMV}~\cite{baraldi2018} trains a video representation using a generated dataset while relying on kernel-based temporal alignment. \textbf{TCA}$_f$~\cite{shao2021} is a transformer-based architecture trained with supervised contrastive learning. \textbf{VRL}~\cite{he2022} is a CNN and transformer-based network trained end-to-end with no labeled data. \textbf{ViSiL}$_f$~\cite{kordopatis2019b} is a baseline without any training on videos that corresponds to the frame-to-frame similarity part of ViSiL combined with Chamfer similarity. \textbf{ViSiL}$_v$ is the full similarity model trained with supervision. \textbf{DnS}~\cite{kordopatis2022} is a ViSiL-based student network trained with distillation from a teacher trained with supervision; we compare with the best-performing fine-grained attention student $\textbf{S}^f_\mathcal{A}$. For TCA and VRL, the reported results are taken from the original papers. For the remaining approaches, we run the provided pretrained networks, and following \textbf{DnS}~\cite{kordopatis2022}, we implement LAMV and DML with the same features provided in the official repository\footnote{\url{https://github.com/mever-team/distill-and-select}}. 
        
\begin{figure}[t]
            \centering
            \input{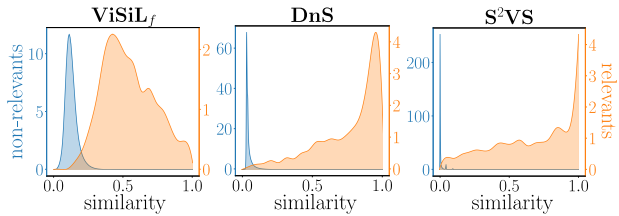}           
            \caption{Similarity distribution of S$^2$VS and ViSiL$_f$ and DnS competitors on the DSVD set of FIVR-200K. For S$^2$VS and DnS the similarities are rescaled to $[0,1]$.
            \vspace{-10pt}        
            \label{fig:sim_dist}}
        \end{figure}

        Table~\ref{tab:mAP_comparison} presents the performance comparison on video retrieval and detection. S$^2$VS is among the top two performing methods in all cases despite not requiring labels. This holds for both training sets used for our method. The best-performing competitor is DnS, which requires a manually labeled dataset of several thousand copied video segments to train a teacher network. Compared to the baseline ViSiL$_f$, S$^2$VS consistently improves performance across all cases by a noticeable margin. The performance improvement is larger for detection due to better similarity calibration across queries, which is demonstrated in Figure~\ref{fig:sim_dist}. S$^2$VS achieves the best separation between relevant and non-relevant samples, compared with ViSiL$_f$ and DnS, with the two distributions not significantly overlapping.
			
		In addition, we evaluate our proposed approach on the CC\_WEB\_VIDEO~\cite{wu2007} dataset, but we do not provide detailed results since the achieved performance is saturated. The mAP performance is 98.6\% / 97.5\% /  99.6\% / 99.5\% when trained with DnS-100K for the different versions as listed in DnS paper~\cite{kordopatis2022}. They are on par or slightly better than the competing methods.
        
        \textbf{Making FIVR harder:}
        FIVR-200K contains a large number of easy examples that dominate the estimation of average performance, giving the impression that the task is nearly solved due to the very high and almost saturated performance. To mitigate this, we discard such easy examples from the dataset to make it more challenging. In particular, we remove all database videos that, as duplicate to a query, are ranked in a position of perfect precision, \ie, before all negatives, by ViSiL$_{f}$, DNS, and S$^2$VS. This process results in 4,828 videos identified as easy database examples, which is almost 40\% of the total videos labeled as relevant to any query. We denote the harder version as FIVR-200K$^\mathcal{H}$.
        
        Table~\ref{tab:hard_fivr} presents results for the harder version and the differences compared to the original version in Table~\ref{tab:mAP_comparison}. Compared with the initial results, the performance significantly drops, with the difference ranging from $\approx$25\% up to 36\% of mAP for retrieval and $\approx$30\% up to 50\% of $\mu$AP for detection. Our approach achieves very similar performance to DnS in the three retrieval tasks with a difference of about 1.3-1.8\%, but with a substantial difference of up to 12\% from the baseline. The performance gap is much larger in the case of detection. The proposed method surpasses the other two by a clear margin of more than 15\% from the second best for all tasks, highlighting its effectiveness even in challenging settings.

        \begin{table}[t]
            \centering
            \vspace{-4pt}
            \def\arraystretch{1.1}
\setlength\tabcolsep{2pt}
\scalebox{0.85}{
    \begin{tabular}{l c c c c c c }
        & \multicolumn{3}{c}{\textbf{Retrieval}} & \multicolumn{3}{c}{\textbf{Detection}} \\ \cmidrule(lr){2-4} \cmidrule(lr){5-7} 
        \textbf{Appr.} & \textbf{DSVR} & \textbf{CSVR} & \textbf{ISVR} & \textbf{DSVD} & \textbf{CSVD} & \textbf{ISVD} \\ \midrule
        \textbf{ViSiL$_f$}~\cite{kordopatis2019b}    & 52.3{\color{red} \textsubscript{-36.6}} & 49.4{\color{red} \textsubscript{-35.4}} & 43.3{\color{red} \textsubscript{-28.9}} & 16.0{\color{red} \textsubscript{-50.9}} & 14.0{\color{red} \textsubscript{-45.5}} & 11.1{\color{red} \textsubscript{-34.8}} \\
        \textbf{DnS}~\cite{kordopatis2022}           & 63.1{\color{red} \textsubscript{-29.0}} & 58.8{\color{red} \textsubscript{-28.7}} & \textbf{48.4}{\color{red} \textsubscript{-25.7}} & 34.5{\color{red} \textsubscript{-45.2}} & 25.3{\color{red} \textsubscript{-44.2}} & 19.6{\color{red} \textsubscript{-34.6}}\\
        \textbf{S$^2$VS} (Ours)                      & \textbf{64.4}{\color{red} \textsubscript{-28.1}} & \textbf{60.0}{\color{red} \textsubscript{-27.8}} & 47.1{\color{red} \textsubscript{-26.8}} & \textbf{52.5}{\color{red} \textsubscript{-36.7}} & \textbf{42.6}{\color{red} \textsubscript{-37.7}} & \textbf{35.0}{\color{red} \textsubscript{-29.9}}
    \end{tabular}
}
            \caption{Retrieval mAP (\%) and detection $\mu$AP (\%) comparison on FIVR-200K$^\mathcal{H}$, which is harder than the original FIVR-200K due to easy video removal. \textbf{Bold} indicates the best approach. Red subscripts indicate the performance drop in comparison to the original FIVR-200K.
            \vspace{-15pt}
            \label{tab:hard_fivr}}
        \end{table}
            
\begin{figure}[t]
            \centering
            \input{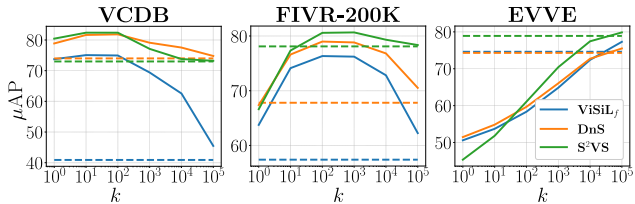}
            \vspace{-15pt} 
            \caption{Detection performance measured via $\mu AP$ with similarity normalization for varying values of $k$ for top-$k$ neighbors. The average performance over the three subtasks is reported on FIVR-200K. Dashed lines indicate performance without normalization.
            \label{fig:score_norm}
            \vspace{-15pt}}
        \end{figure}

        \textbf{Similarity normalization:} To delve further into the detection performance comparison, we apply similarity normalization~\cite{pizzi2022}, initially proposed for image copy detection, on top of all approaches. For each query video, the top-$k$ neighbors in a background set are estimated. Then, their similarity to the query is averaged to form the query's bias term, which is subtracted from the initial similarities. This impacts the global ranking of video pairs, as each query in the dataset has a different bias term. In this experiment, we use the DnS-100K as the background set and calculate the bias term based on the average similarity of the top-$k$ neighbors. Figure~\ref{fig:score_norm} presents the results with and without similarity normalization. All methods benefit from a wide range of $k$ on VCDB and FIVR-200K, but only for very large values on EVVE, while the baseline benefits the most as there is more space for improvements. Our method achieves the best performance after normalization too. Nevertheless, $k$ needs to be tuned independently per test set, which is a major drawback. Therefore, the good detection performance of our approach, even without normalization, is important.
        \looseness=-1
     
    \subsection{Ablation study}\label{sec:ablation_study}
		
        We perform ablations for different augmentation types and loss functions. For further ablations, we point readers to the supplementary material.
		
        \smallskip\noindent\textbf{Impact of augmentations}: Table \ref{tab:augmentations} presents the performance of the proposed approach for different training augmentation strategies. Each newly added augmentation gives a performance boost, while performance improves by a large margin compared to no use of strong augmentations. We conclude that the variety and diversity of the augmentations are the key ingredients for high performance. \looseness=-1
        
        \smallskip\noindent\textbf{Impact of $\mathcal{L}_{sshn}$ loss}: Table \ref{tab:losses} reports the mAP and $\mu$AP of our method trained with and without the proposed $\mathcal{L}_{sshn}$ loss. It is evident that without the use of the proposed loss, the network does not work effectively. In all evaluation cases, the performance difference is a least 3\%, with the most notable discrepancies on detection runs where it reaches almost 10\% on VCDB. This highlights that the proposed loss is necessary for the effective training of such a video similarity network.

        \smallskip\noindent\textbf{Label supervision vs self-supervision}: Table \ref{tab:super_vs_ssl} presents the results of our approach while positives and negatives are drawn from VCSL, which is an annotated dataset. For this supervised variant, the same losses and augmentations are used except for the video-in-video because we found several such cases in the annotated pairs of VCSL. During batch construction, we make sure that the selected video segments for the positive videos overlap with the dataset ground truth. With our SSL training scheme, the model achieves significantly better results, especially for detection, where it outperforms the supervised counterpart by a margin of up to 9\%. This highlights the need for a large and diverse dataset to effectively train these video similarity learning models.

        \begin{table}[t]
          \centering
          \def\arraystretch{1.1}
\setlength\tabcolsep{3pt}
\scalebox{0.85}{
    \begin{tabular}{l c c c c c c}
        & \multicolumn{3}{c}{\textbf{Retrieval}} & \multicolumn{3}{c}{\textbf{Detection}} \\ \cmidrule(lr){2-4} \cmidrule(lr){5-7}
        \textbf{Augmentations} & \textbf{VCDB} & \textbf{FIVR} & \textbf{EVVE} & \textbf{VCDB} & \textbf{FIVR} & \textbf{EVVE} \\ \midrule
        no strong aug.      & 88.8  & 78.5 & 50.6 & 79.8 & 66.9 & 65.5  \\
        + global trans.     & 94.3  & 83.4 & 59.5 & 89.5 & 74.7 & 76.3  \\
        + frame trans.      & 95.1  & 86.2 & 64.1 & 90.0 & 79.8 & 76.1  \\
        + temporal trans.   & 95.2  & 86.5 & 64.8 & 89.9 & 80.8 & 76.8  \\ 
        + video-in-video    & 95.2  & 87.0 & 65.9 & 90.1 & 81.7 & 78.9  \\
    \end{tabular}
}
          \vspace{-5pt}
          \caption{Retrieval mAP (\%) and detection $\mu$AP (\%) of S$^2$VS with different augmentation strategies.
          \vspace{-5pt}
          \label{tab:augmentations}}
        \end{table}
        
        \begin{table}[t]
          \centering
          \def\arraystretch{1.1}
\setlength\tabcolsep{3pt}
\begin{tabular}{c c c c c c c}
    & \multicolumn{3}{c}{\textbf{Retrieval}} & \multicolumn{3}{c}{\textbf{Detection}} \\ \cmidrule(lr){2-4} \cmidrule(lr){5-7} 
    $\mathcal{L}_{\text{sshn}}$ & \textbf{VCDB} & \textbf{FIVR} & \textbf{EVVE} & \textbf{VCDB} & \textbf{FIVR} & \textbf{EVVE} \\ \midrule
    \ding{55}     & 89.2  & 81.6 & 62.6 & 80.3 & 72.3 & 75.4  \\
    \checkmark    & 95.2  & 87.0 & 65.9 & 90.1 & 81.7 & 78.9  \\
\end{tabular}
          \vspace{-5pt}
          \caption{Retrieval mAP (\%) and detection $\mu$AP (\%) of S$^2$VS with and without $\mathcal{L}_{sshn}$ loss.
          \vspace{-5pt}
          \label{tab:losses}}
        \end{table}
            
        \begin{table}[t]
          \centering
          \setlength\tabcolsep{2.5pt}
\scalebox{0.85}{
\begin{tabular}{l l c c c c c c}
    & & \multicolumn{3}{c}{\textbf{Retrieval}} & \multicolumn{3}{c}{\textbf{Detection}} \\ \cmidrule(lr){3-5} \cmidrule(lr){6-8} 
    \textbf{Train} & \textbf{Dataset} & \textbf{VCDB} & \textbf{FIVR} & \textbf{EVVE} & \textbf{VCDB} & \textbf{FIVR} & \textbf{EVVE} \\ \midrule
    Sup. & VCSL                 & 92.4 & 83.8 & 64.0 & 85.6 & 72.7 & 73.5  \\
    SSL  & DnS-100K             & 95.2 & 87.0 & 65.9 & 90.1 & 81.7 & 78.9  \\
    SSL  & VCDB ($\mathcal{D}$) & -    & 87.3 & 67.2 & -    & 78.1 & 80.7  \\
\end{tabular}
}
          \vspace{-5pt}
          \caption{Retrieval mAP (\%) and detection $\mu$AP (\%) for the proposed method with self-supervision on two different datasets and a variant that uses labeled positives and negatives.
          \vspace{-7pt}
          \label{tab:super_vs_ssl}}
        \end{table}

\section{Conclusions}
\label{sec:conclusions}
In this paper, we propose a self-supervised learning approach for training video similarity networks. Eliminating the need for labels allows us to train on large-scale video corpora, which, together with a diverse set of video augmentations, form the key ingredient for achieving top performance. The obtained single model is evaluated on several target retrieval and detection tasks. It manages to perform on par or outperform existing models that exploit labeled datasets, especially for detection due to better similarity calibration across queries.

\bigskip\noindent\textbf{Acknowledgments:}
This work was supported by Junior Star GACR under grant No. GM 21-28830M, and the EU partially funded projects H2020 MediaVerse, H2020 AI4Media and Horizon Europe vera.ai under contract No. 957252, 951911 and 101070093, respectively.

\newpage

\begin{center}
\Large \textbf{Supplementary materials}
\vspace{10pt}
\end{center}

\renewcommand\thesection{\Alph{section}}

\setcounter{section}{0}

\section{Additional ablations}
    In this section, we continue our ablations studying how various hyperparameters of the training processes affect the final performance of the proposed method. We conduct further experiments on the same datasets as in the main paper.

    \smallskip\noindent\textbf{Impact of $\lambda$ hyperparameter}: In Table~\ref{tab:lambda}, we report the results of the proposed approach for different values of $\lambda$. We observe that the performance is not significantly affected for smaller than the default $\lambda=3$ values. However, for larger values, it steadily decreases as the network focuses more on the SSHN loss than the InfoNCE one.

    \smallskip\noindent\textbf{Impact of temperature $\tau$}: Table~\ref{tab:tau} presents the performance of S$^2$VS trained with different temperature values in the InfoNCE loss. The performance decreases for values other than the default $\tau=0.03$, which is more noticeable on detection tasks where larger $\mu$AP drops are reported. We do not go lower due to numeric instability and overflow issues during training, causing our network to collapse. 

    \smallskip\noindent\textbf{Impact of the number of $T_B$ frames}: Table~\ref{tab:frame_num} displays the results of our method trained with different number of $T_B$ frames for the input videos. In almost all tasks, the larger the size of input videos, the better the performance. Also, when $T_B$ is small, the network fails to learn anything useful. This is expected as comparing larger videos helps the model better capture temporal structures in the similarity matrices.

    \smallskip\noindent\textbf{Impact of $\mathcal{L}_{\text{sshn}}$ terms}: Table~\ref{tab:sshn_terms} illustrates the results of the proposed approach trained with the different terms of $\mathcal{L}_{\text{sshn}}$ loss, with $\mathcal{L}_{\text{ss}}$ and $\mathcal{L}_{\text{hn}}$ standing for the self-similarity and hard negative part of the loss (cf. Equation 3, in the main paper). This highlights that both terms are necessary for the effective training of the system.
        
     \smallskip\noindent\textbf{Mean and standard deviation}: Table~\ref{tab:mean_std} shows the mean and standard deviation of mAP and $\mu$AP for retrieval and detection, respectively, of S$^2$VS trained and evaluated seven times with different seeds. Generally, the performance is steady with small fluctuations, especially for VCDB, where the standard deviation is less than 0.1\%, and with the largest deviations reported on EVVE.

\section{Comparison per video query}
    
\begin{figure*}
        \centering
        \begin{subfigure}{\textwidth}
  \centering
  \begin{tabular}{ccc}
    \includegraphics[width=.29\textwidth]{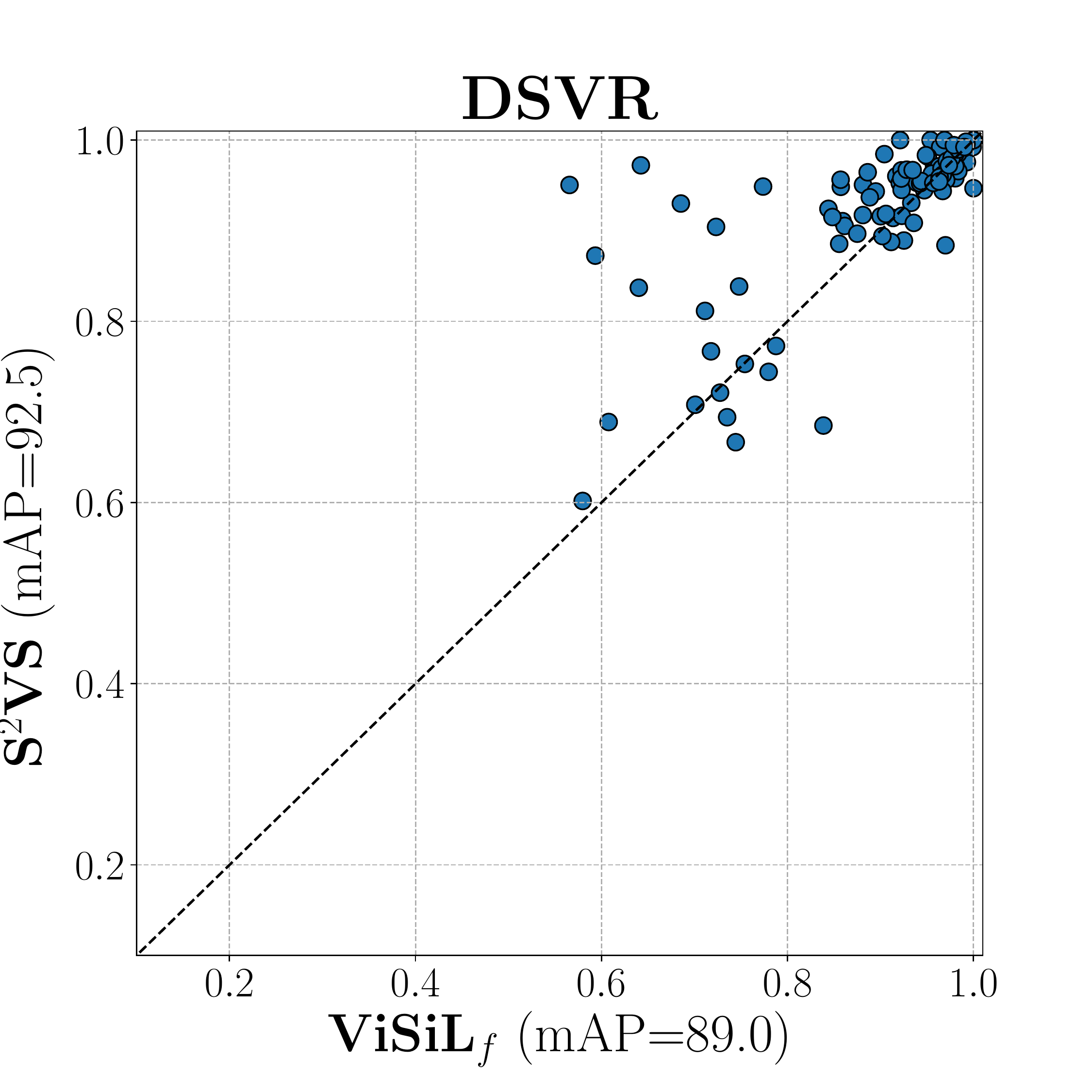} &
    \includegraphics[width=.29\textwidth]{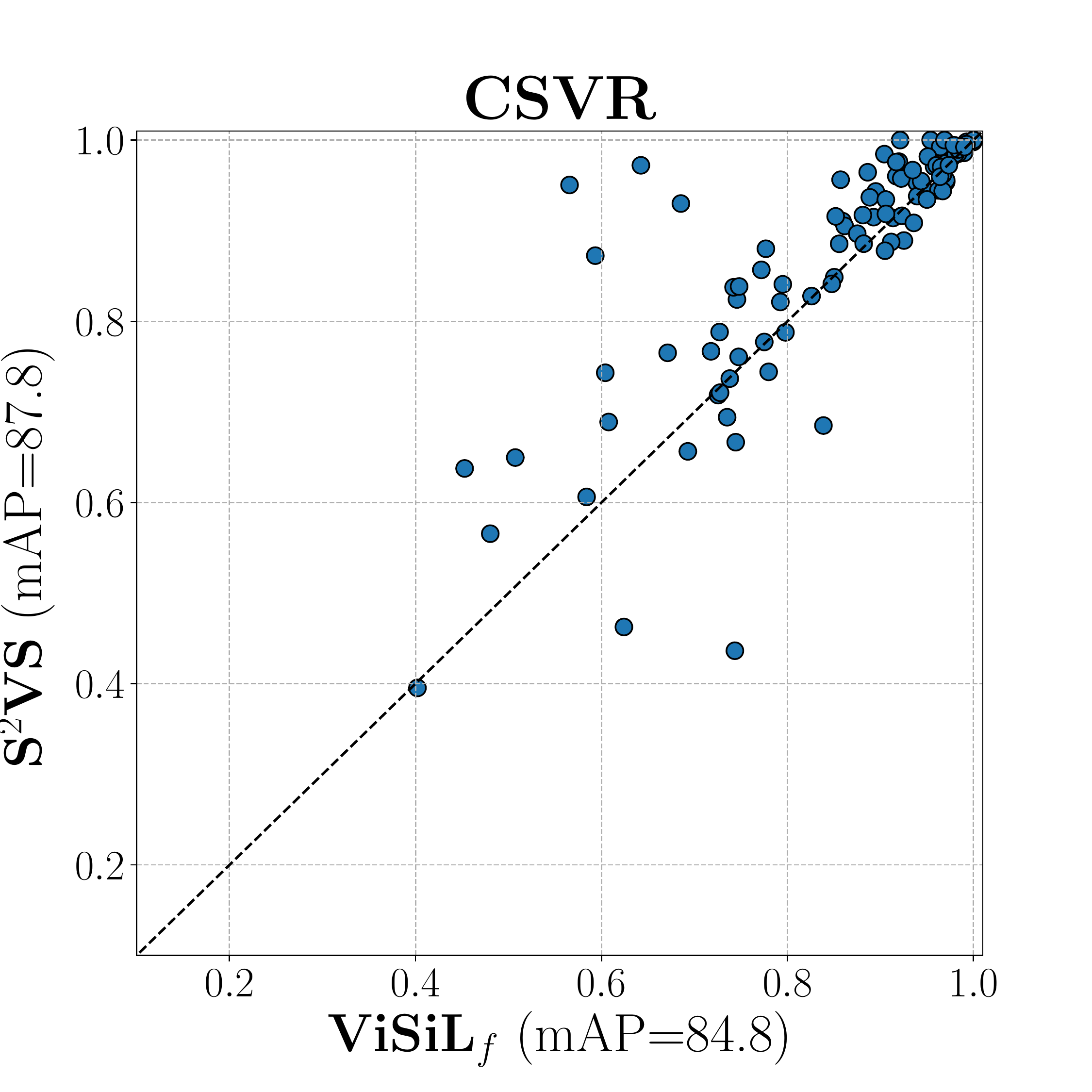} &
    \includegraphics[width=.29\textwidth]{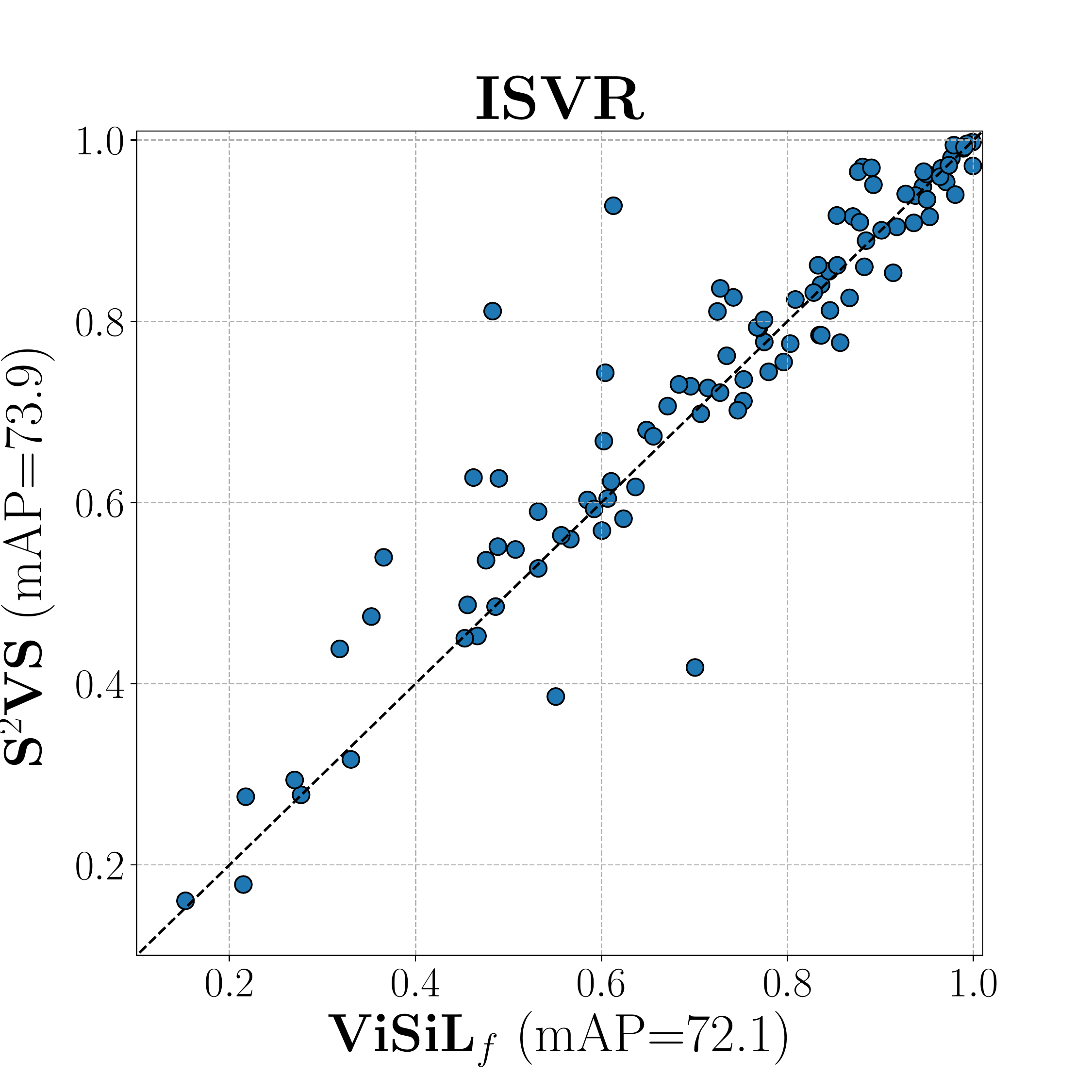}  \\
    \includegraphics[width=.29\textwidth]{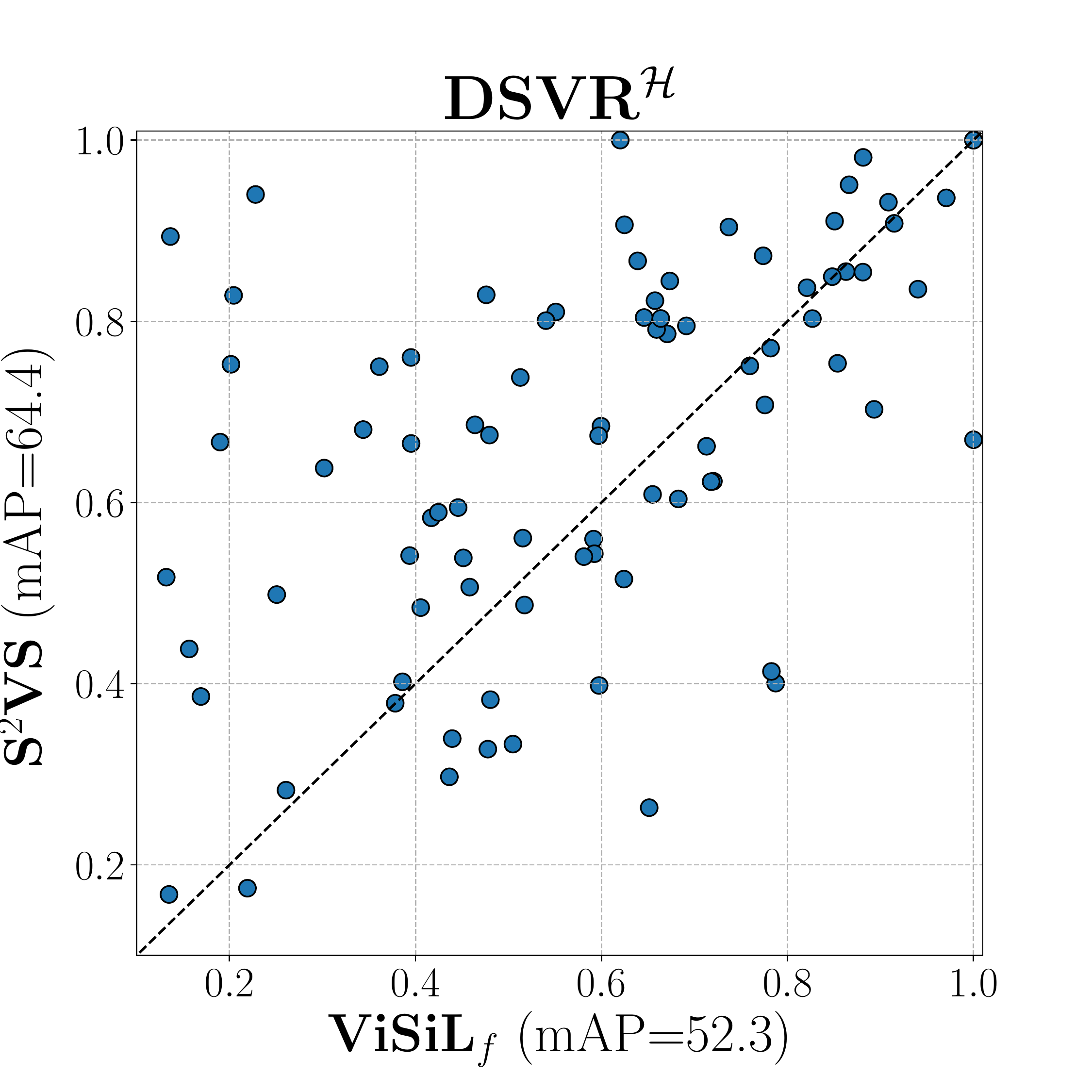} &
    \includegraphics[width=.29\textwidth]{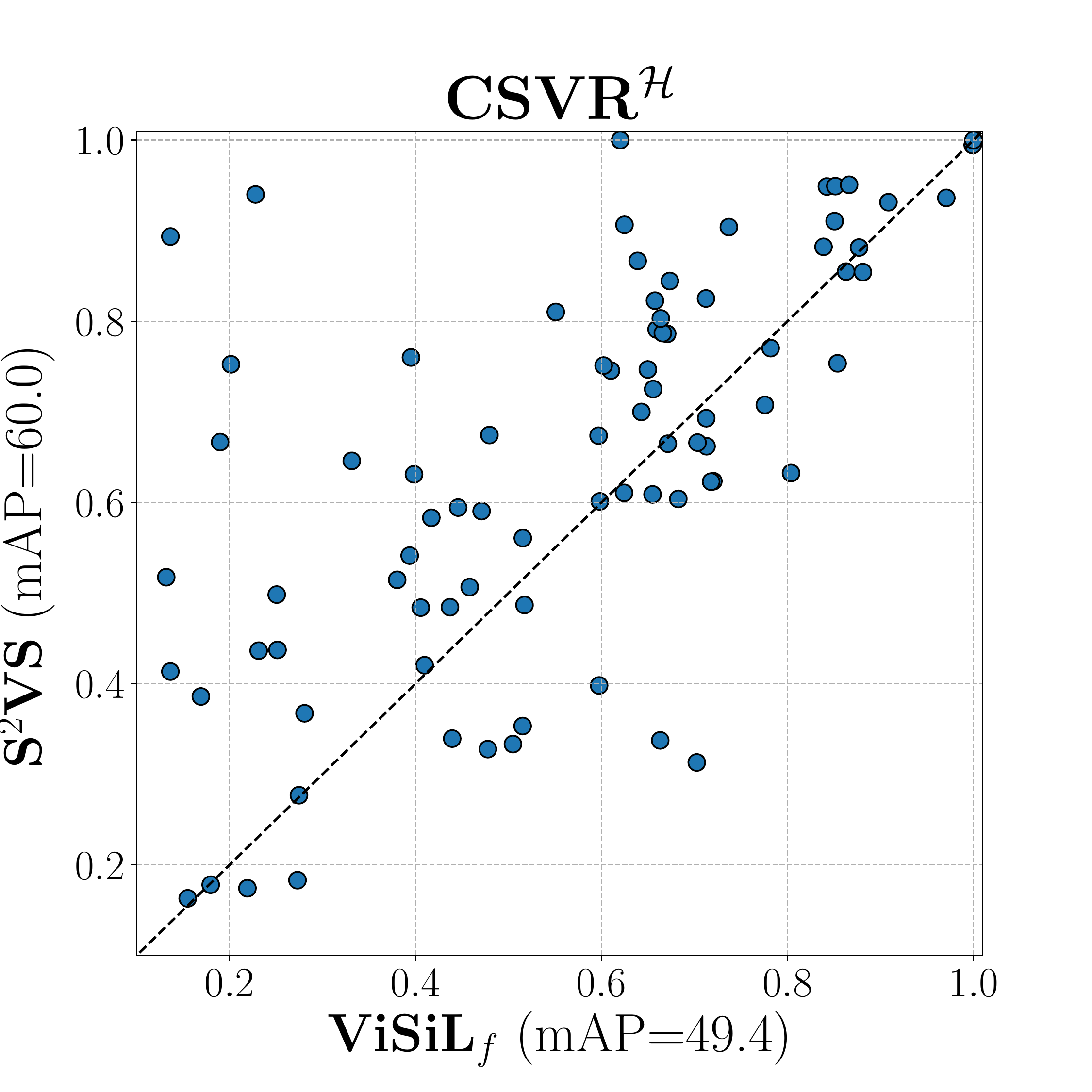} &
    \includegraphics[width=.29\textwidth]{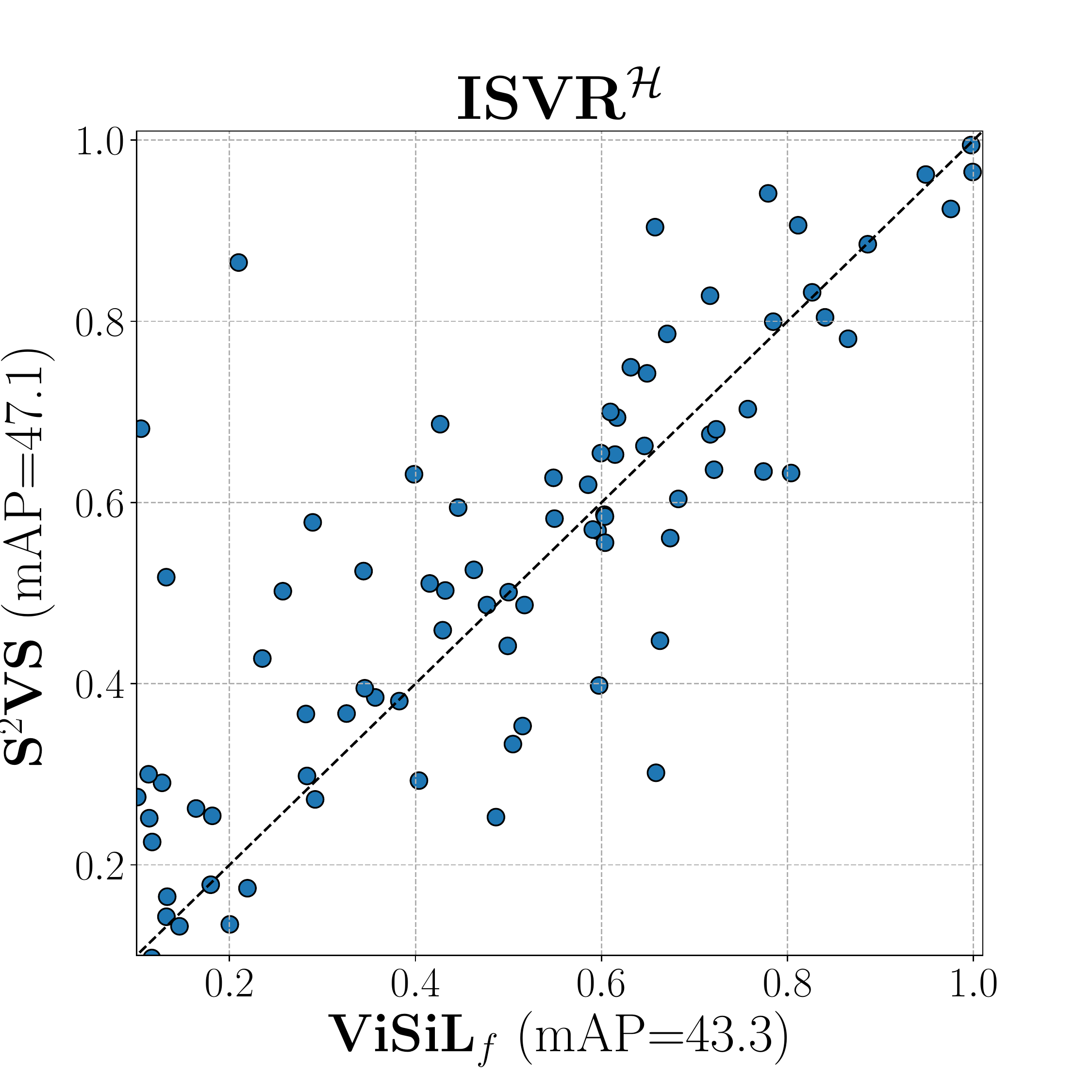} 
  \end{tabular}
  \caption{ViSiL$_{f}$ -- S$^2$VS}
\end{subfigure}

\begin{subfigure}{\textwidth}
  \centering
  \begin{tabular}{ccc}
    \includegraphics[width=.29\textwidth]{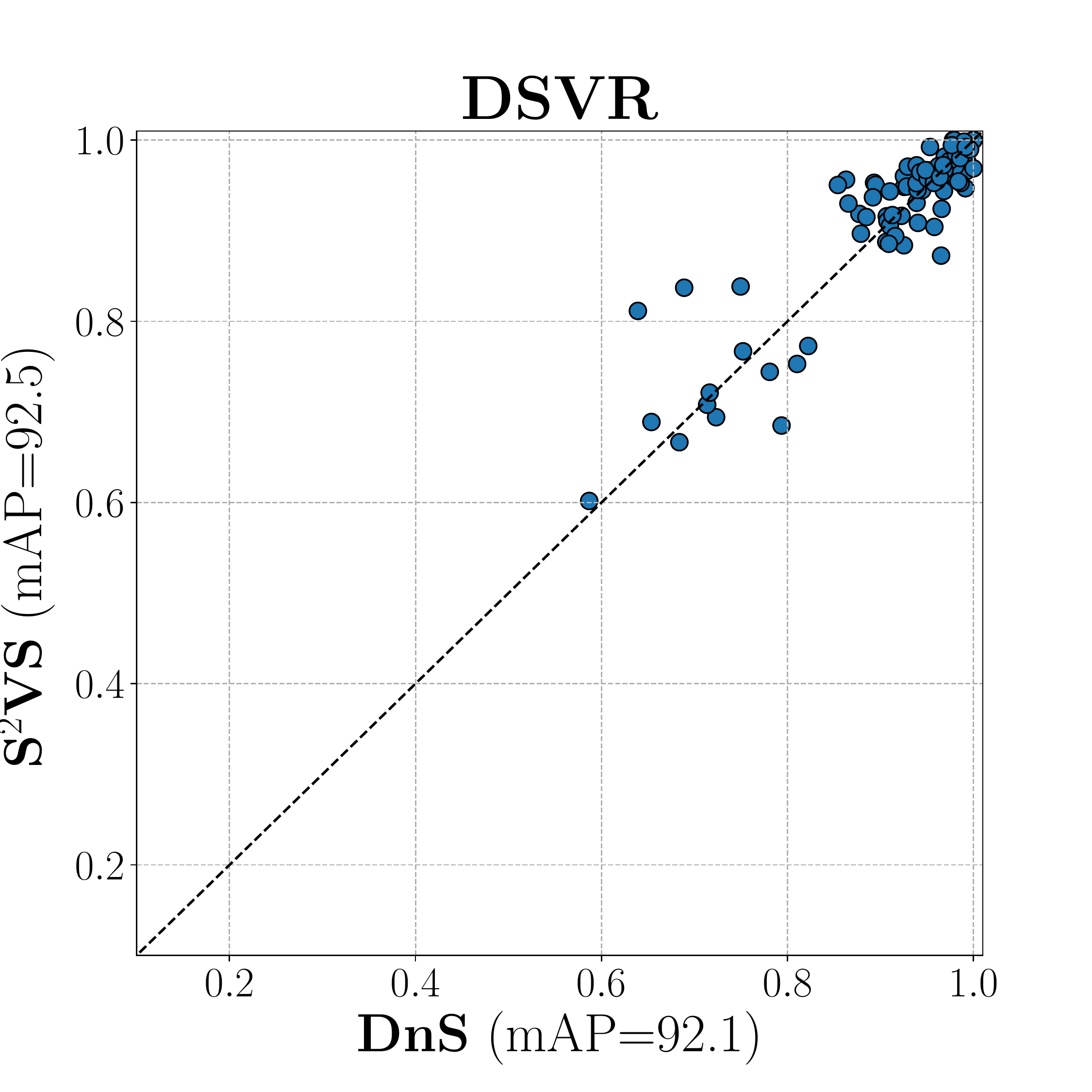} &
    \includegraphics[width=.29\textwidth]{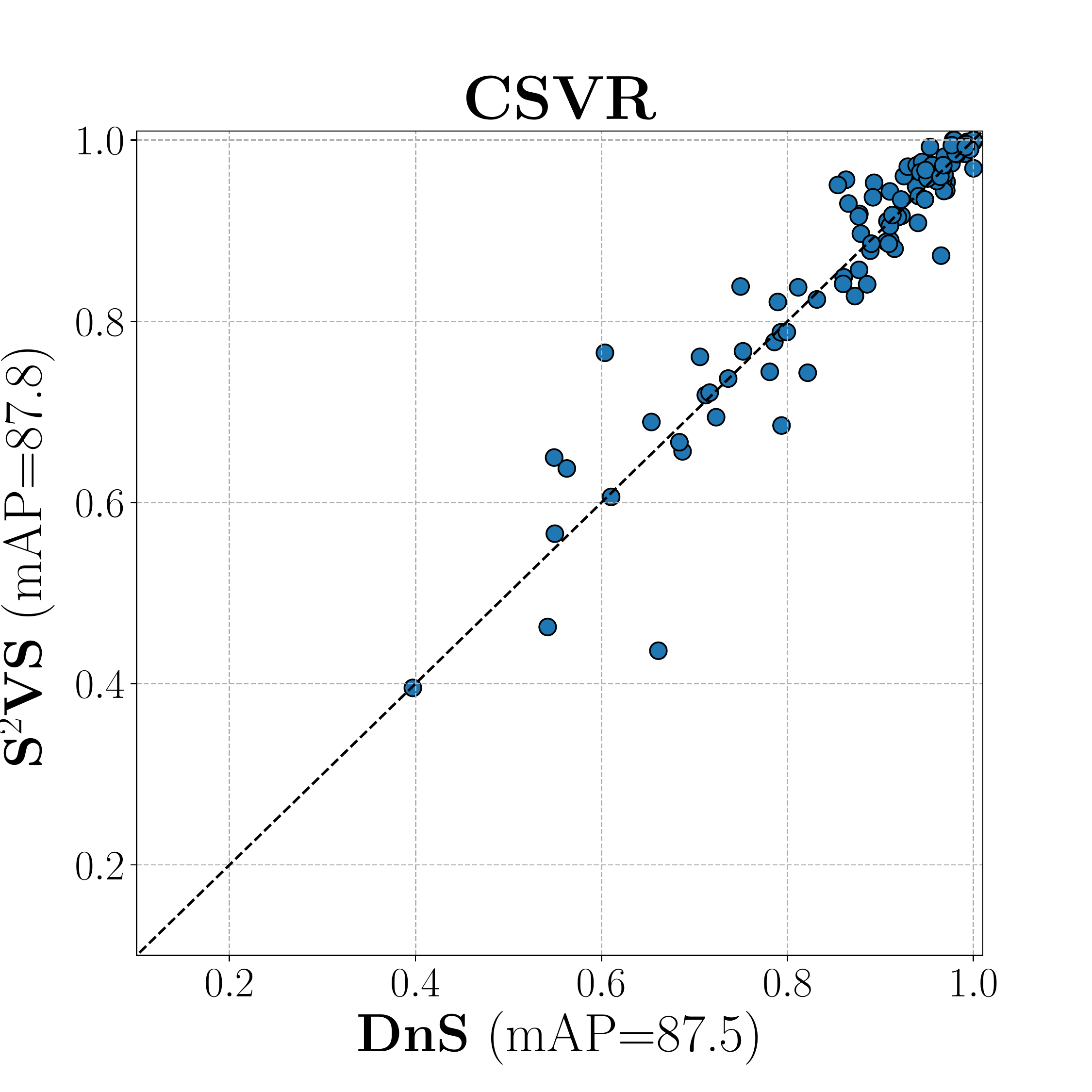} &
    \includegraphics[width=.29\textwidth]{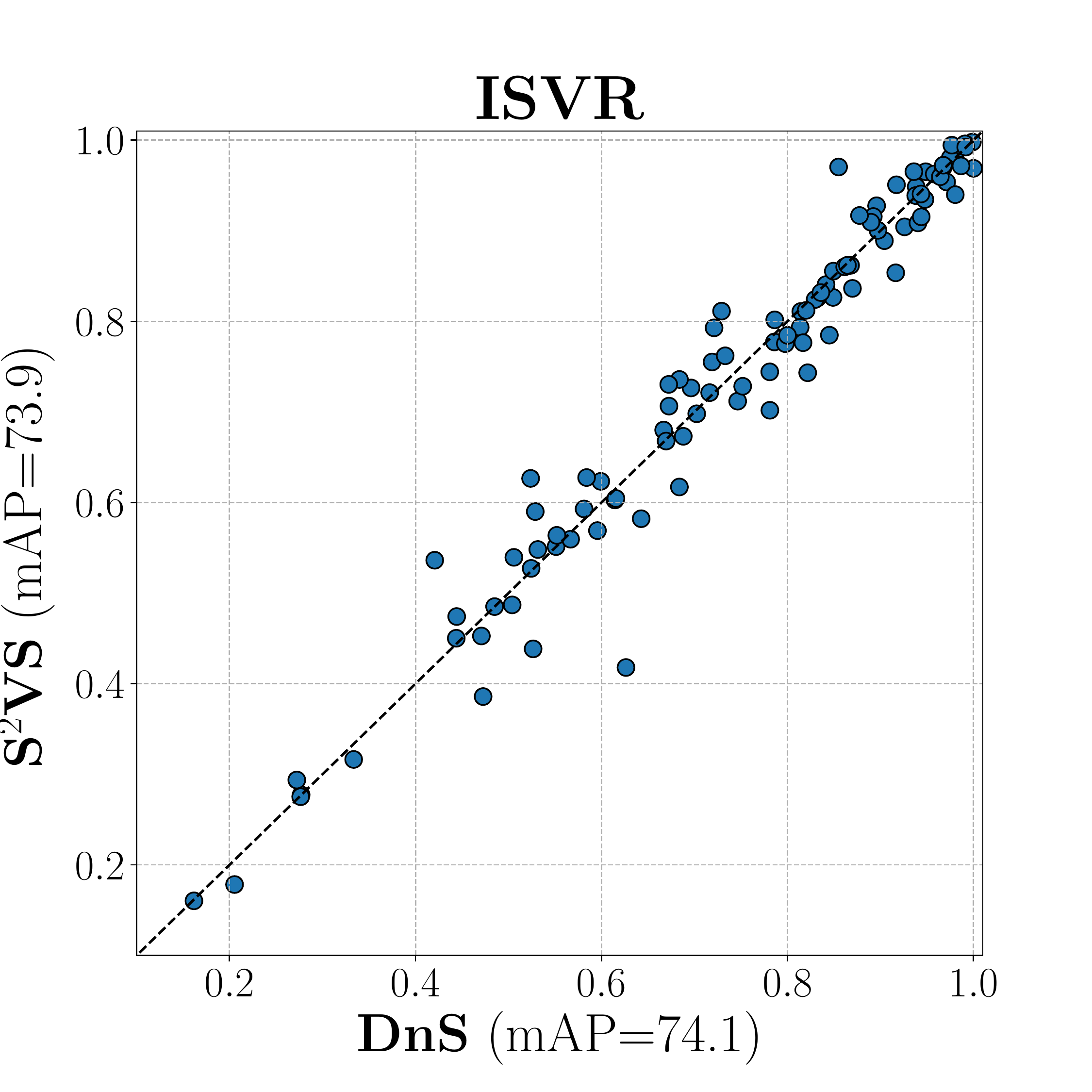}  \\
    \includegraphics[width=.29\textwidth]{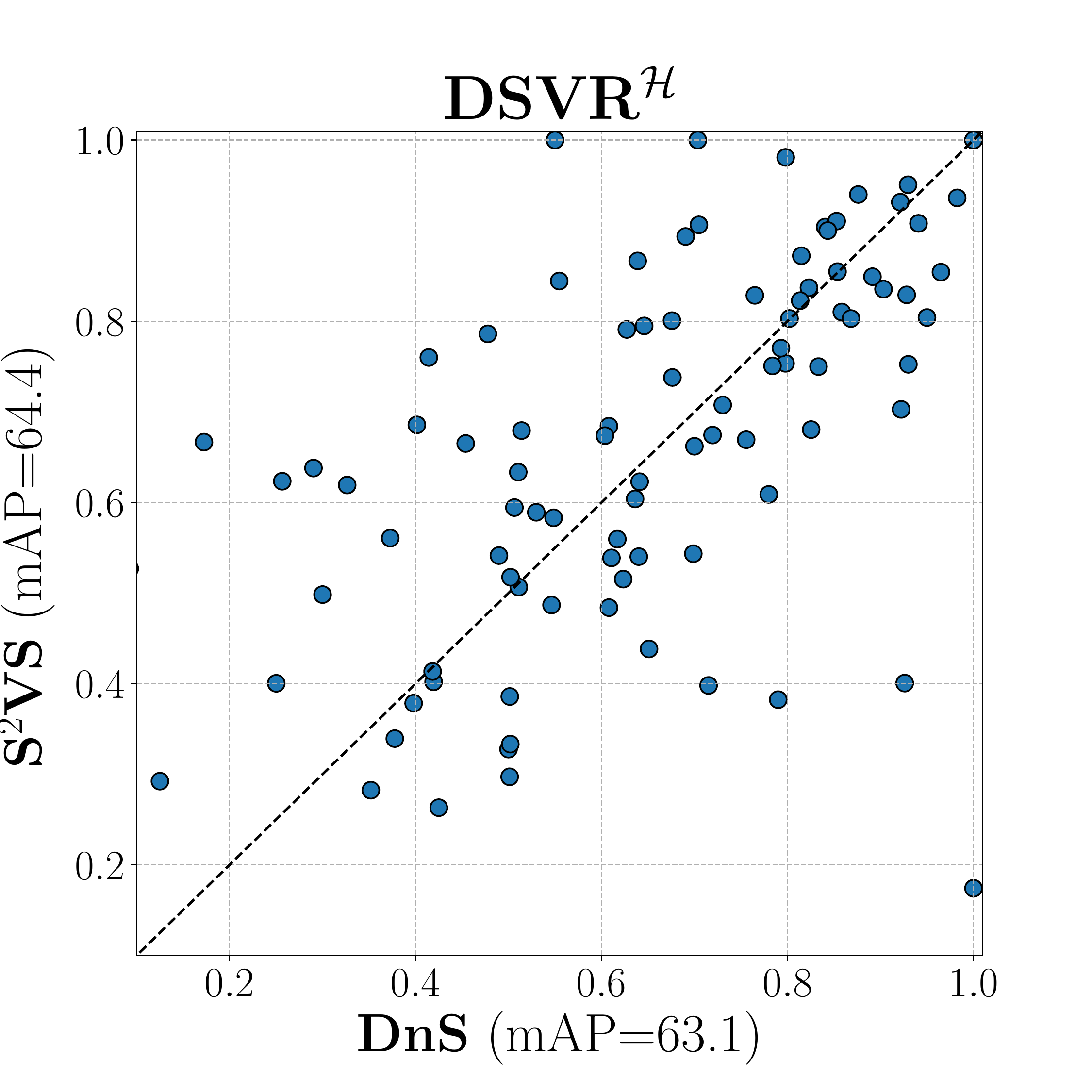} &
    \includegraphics[width=.29\textwidth]{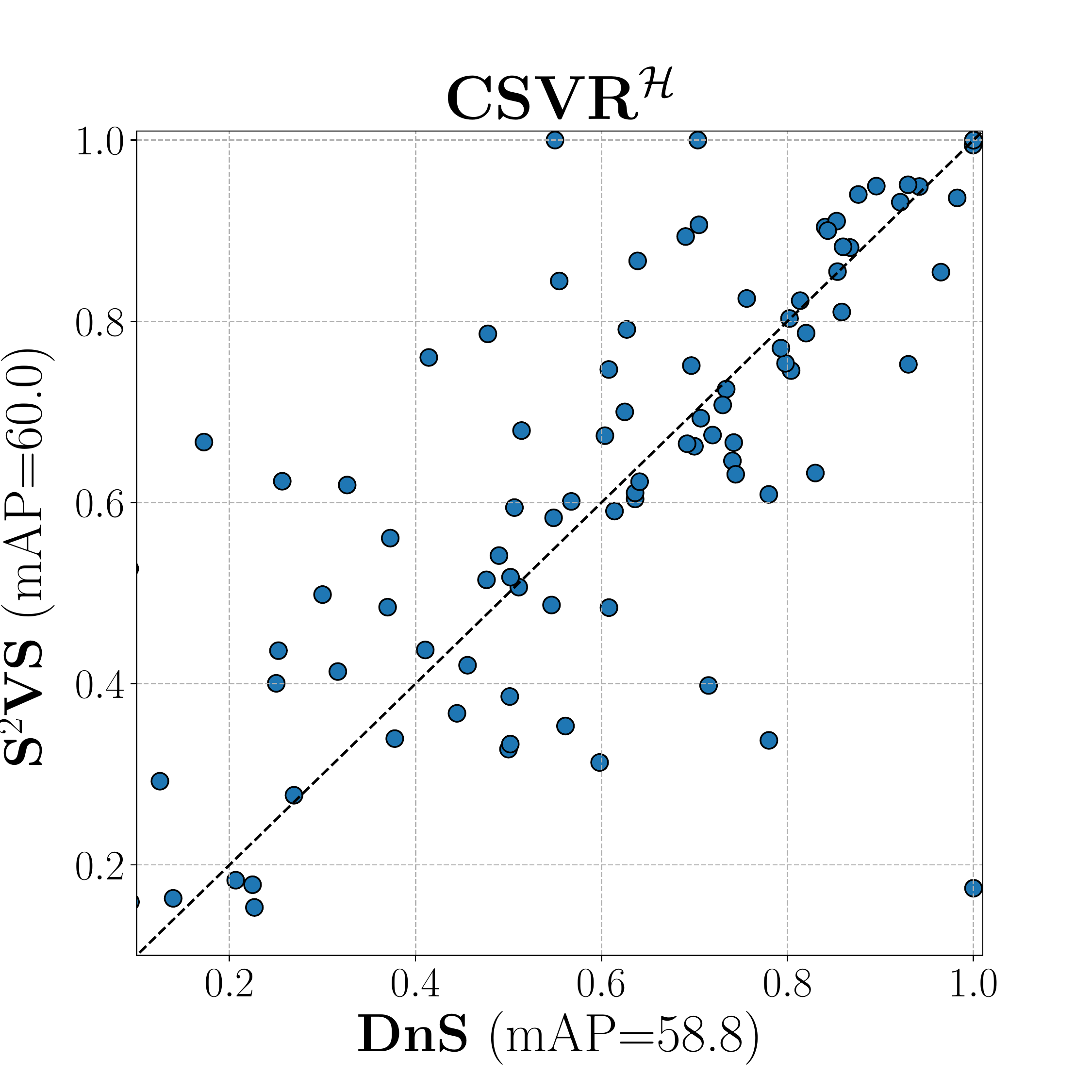} &
    \includegraphics[width=.29\textwidth]{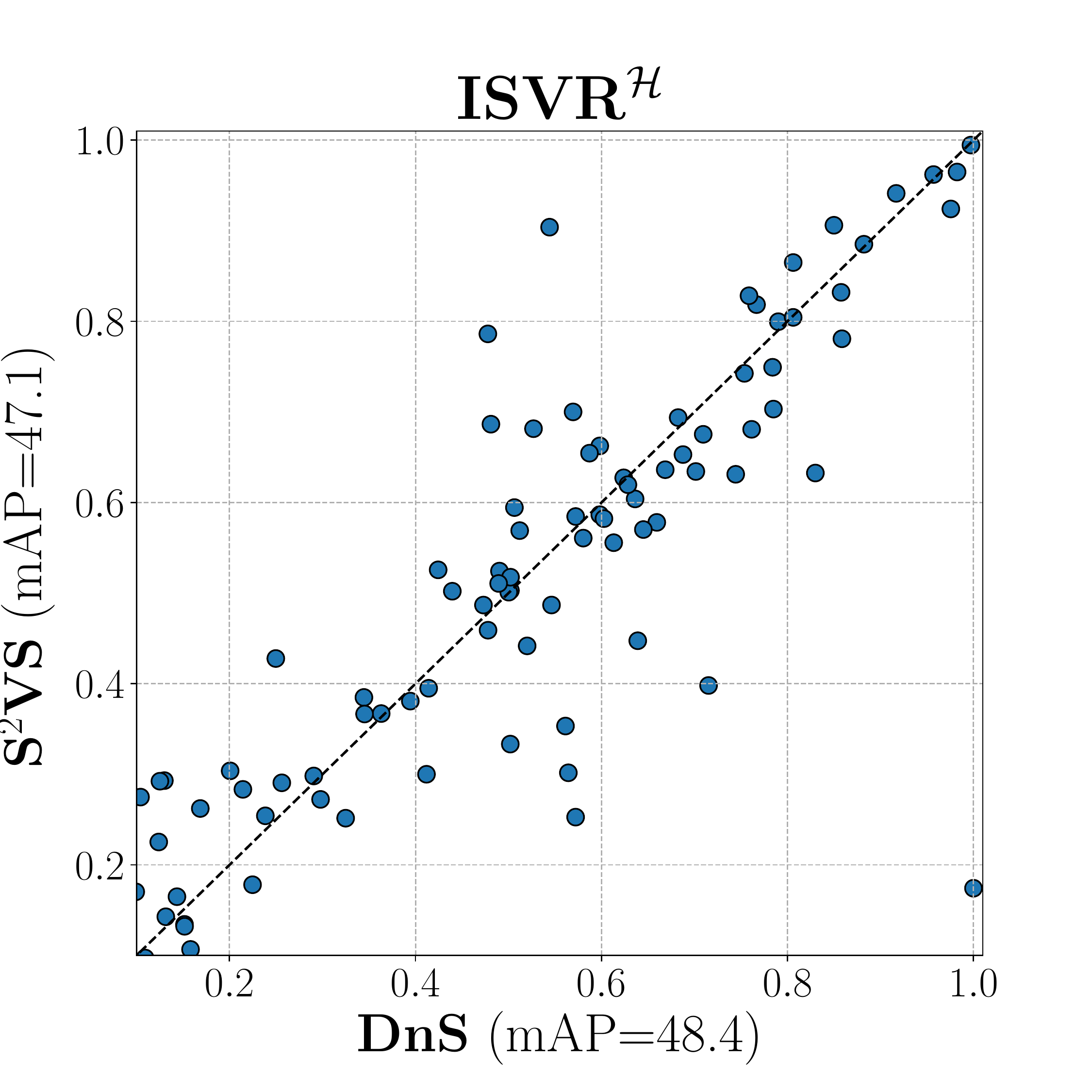}
  \end{tabular}
  \caption{DnS -- S$^2$VS}
\end{subfigure}
        \vspace{-10pt}
        \caption{Average Precision (AP) per query for the proposed S$^2$VS in comparison with the \textbf{ViSiL$_f$}~\cite{kordopatis2019b}, and \textbf{DnS}~\cite{kordopatis2022} on the three subtasks of FIVR-200K and on its hard subset FIVR-200K$^\mathcal{H}$ (cf. Section 5.1, in the main paper).
        \label{fig:scatter}
        }
    \end{figure*}

    In this section, we compare the per query performance of the proposed approach with \textbf{ViSiL$_f$}~\cite{kordopatis2019b} and \textbf{DnS}~\cite{kordopatis2022}. Figure~\ref{fig:scatter} illustrates our method's Average Precision (AP) for each query on the FIVR-200K dataset and its hard version in comparison to ViSiL$_f$ and DnS. The diagonal line indicates the cases where there is a tie performance between the two compared approaches. In the normal settings of the dataset, comparing S$^2$VS with ViSiL$_f$, most of the queries lie on the part above the diagonal line for all three tasks, indicating that the proposed method achieves better results on them, while a large number of points appear on the top-right corner indicating easy queries for all approaches. Additionally, comparing S$^2$VS with DnS, more queries are close to diagonal, which means that the two approaches have similar performance. Nevertheless, it is noteworthy that the vast majority of the queries are very close to one, dominating the results and making differences in performance less apparent in the final evaluation.
    
        \begin{table}[t]
          \centering
          \setlength\tabcolsep{5pt}
\scalebox{0.99}{
    \begin{tabular}{l c c c c c c}
        & \multicolumn{3}{c}{\textbf{Retrieval}} & \multicolumn{3}{c}{\textbf{Detection}} \\ \cmidrule(lr){2-4} \cmidrule(lr){5-7} 
        $\lambda$\quad & \textbf{VCDB} & \textbf{FIVR} & \textbf{EVVE} & \textbf{VCDB} & \textbf{FIVR} & \textbf{EVVE} \\ \midrule
        \textit{0}   & 89.2 & 81.6 & 62.6 & 80.3 & 72.3 & 75.4  \\
        \textit{1}   & 94.1 & 85.9 & 64.7 & 88.2 & 80.6 & 78.5  \\ 
        \textit{3}   & 95.2 & 87.0 & 65.9 & 90.1 & 81.7 & 78.9  \\
        \textit{5}   & 95.3 & 86.3 & 65.2 & 90.5 & 81.3 & 78.1  \\ 
        \textit{7}   & 94.6 & 85.5 & 64.5 & 89.2 & 79.2 & 78.1  \\ 
    \end{tabular}
}
          \vspace{-8pt}
          \caption{Retrieval mAP (\%) and detection $\mu$AP (\%) for S$^2$VS with different values for scale factor $\lambda$.
          \vspace{-6pt}
          \label{tab:lambda}}
        \end{table}
    
        \begin{table}[t]
          \centering
          \setlength\tabcolsep{3.5pt}
\begin{tabular}{l c c c c c c}
    & \multicolumn{3}{c}{\textbf{Retrieval}} & \multicolumn{3}{c}{\textbf{Detection}} \\ \cmidrule(lr){2-4} \cmidrule(lr){5-7} 
    $\tau$\quad & \textbf{VCDB} & \textbf{FIVR} & \textbf{EVVE} & \textbf{VCDB} & \textbf{FIVR} & \textbf{EVVE} \\ \midrule
    \textit{0.03}   & 95.2 & 87.0 & 65.9 & 90.1 & 81.7 & 78.9  \\
    \textit{0.05}   & 95.2 & 86.7 & 65.4 & 89.9 & 80.4 & 77.5  \\ 
    \textit{0.07}   & 95.1 & 86.0 & 65.3 & 89.7 & 78.8 & 76.8  \\
    \textit{0.1}    & 95.1 & 85.7 & 65.0 & 89.5 & 78.0 & 75.5  \\ 
\end{tabular}
          \vspace{-8pt}
          \caption{Retrieval mAP (\%) and detection $\mu$AP (\%) for S$^2$VS with different values for temperature $\tau$.
          \vspace{-6pt}
          \label{tab:tau}}
        \end{table}	
		
        \begin{table}[t]
          \centering
          \setlength\tabcolsep{3.5pt}
\scalebox{0.99}{
    \begin{tabular}{l c c c c c c}
        & \multicolumn{3}{c}{\textbf{Retrieval}} & \multicolumn{3}{c}{\textbf{Detection}} \\ \cmidrule(lr){2-4} \cmidrule(lr){5-7} 
        $T_B$\quad & \textbf{VCDB} & \textbf{FIVR} & \textbf{EVVE} & \textbf{VCDB} & \textbf{FIVR} & \textbf{EVVE} \\ \midrule
        \textit{8}   & 86.8 & 70.1 & 52.6 & 78.5 & 57.0 & 66.7  \\ 
        \textit{16}  & 95.1 & 86.9 & 65.2 & 89.6 & 82.0 & 78.8  \\ 
        \textit{32}  & 95.2 & 87.0 & 65.9 & 90.1 & 81.7 & 78.9  \\
    \end{tabular}
}
          \vspace{-8pt}
          \caption{Retrieval mAP (\%) and detection $\mu$AP (\%) for S$^2$VS with different number of $T_B$ frames.
          \vspace{-6pt}
          \label{tab:frame_num}}
        \end{table}
        
		\begin{table}[t]
          \centering
          \setlength\tabcolsep{3pt}
\begin{tabular}{c c c c c c c c}
    & & \multicolumn{3}{c}{\textbf{Retrieval}}  & \multicolumn{3}{c}{\textbf{Detection}} \\ \cmidrule(lr){3-5} \cmidrule(lr){6-8} 
    $\mathcal{L}_{\text{ss}}$ & $\mathcal{L}_{\text{hn}}$  & \textbf{VCDB}   & \textbf{FIVR}    & \textbf{EVVE} &  \textbf{VCDB} & \textbf{FIVR} & \textbf{EVVE} \\ \midrule
    \ding{55}     & \ding{55}     & 89.2 & 81.6 & 62.6 & 80.3 & 72.3 & 75.4  \\
    \checkmark    & \ding{55}     & 91.9 & 84.0 & 61.8 & 79.5 & 68.1 & 64.5  \\
    \ding{55}     & \checkmark    & 94.5 & 84.8 & 64.6 & 88.1 & 78.9 & 75.1  \\
    \checkmark    & \checkmark    & 95.2 & 87.0 & 65.9 & 90.1 & 81.7 & 78.9  \\
\end{tabular}
          \vspace{-8pt}
          \caption{Retrieval mAP (\%) and detection $\mu$AP (\%) for S$^2$VS trained with different configurations for $\mathcal{L}_{\text{sshn}}$ loss.
          \vspace{-6pt}
          \label{tab:sshn_terms}}
        \end{table}

    	\begin{table}[t]
          \centering
          \setlength\tabcolsep{3pt}
\scalebox{0.99}{
\begin{tabular}{l c c c }
    \textbf{task} & \textbf{VCDB} & \textbf{FIVR} & \textbf{EVVE} \\ \midrule
    retrieval   & 95.2 $\pm$ 0.07 & 87.0 $\pm$ 0.20 & 65.9 $\pm$ 0.18  \\ 
    detection   & 90.0 $\pm$ 0.09 & 81.8 $\pm$ 0.21 & 78.8 $\pm$ 0.51  \\ 
\end{tabular}
}
          \vspace{-8pt}
          \caption{Mean and standard deviation of retrieval mAP (\%) and detection $\mu$AP (\%) of S$^2$VS.
          \vspace{-10pt}
          \label{tab:mean_std}}
		\end{table}
		
		\begin{figure*}[t]
            \centering
            \includegraphics[width=\textwidth]{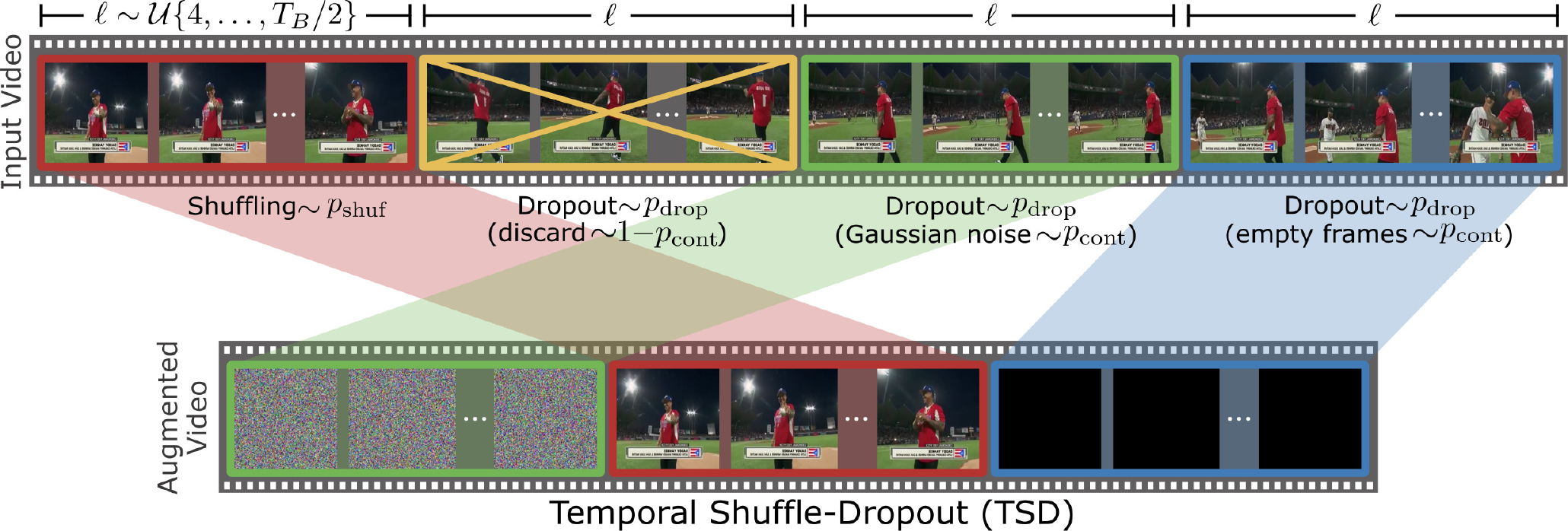}
            \caption{Illustration of the proposed Temporal Shuffle-Dropout (TSD) augmentation scheme, where an input video is split into short clips of fixed length $\ell\sim\mathcal{U}\{4,\cdots,T_B/2\}$, each one of them being shuffled with probability $p_{\text{shuf}}$, or dropped out with probability $p_{\text{drop}}$ (i.e., filled with empty frames or Gaussian noise with probability $p_{\text{cont}}$, or entirely discarded with probability $1-p_{\text{cont}}$). This guarantees the preservation of the local (clip-level) temporal order, but alters the global (video-level) temporal structure.
            \label{fig:tsd}}
        \end{figure*}   
        
    However, in the hard version of the dataset, the query APs are more spread out in the off-diagonal area, highlighting that the methods' performance is less correlated than in the initial settings. In comparison with ViSiL$_f$, more queries lie on the upper left part of the figure, indicating that our S$^2$VS performs better on them. Such queries dominate the mAP and result in a significant performance gap between the two approaches. On the other hand, in comparison with DnS, the queries appear more equally distributed in the off-diagonal areas, resulting in very similar mAP scores. Hence, one method is more effective than the other on a different set of queries. This highlights that there is still much room for improvement, \eg, selecting the appropriate method for the corresponding queries.

\section{Temporal Shuffle Dropout visualization}

    For a better understanding of the proposed Temporal Shuffle Dropout (TSD) transformation scheme, in Figure~\ref{fig:tsd}, we illustrate the various augmentation operations applied in an input video. In this example, the input video is first split into four clips (each being highlighted in a different colour) of $\ell\sim\mathcal{U}\{4,\cdots,T_B/2\}$ frames, as described in  Section~3 of the main paper. Then, a separate operation is applied on each clip, with certain probabilities as shown in Figure~\ref{fig:tsd}, in order to form the augmented video that is then used during the training of our approach: (i) the red one is shuffled, i.e., it is moved between the green and the blue clips (shuffling phase), (ii) the yellow one is discarded entirely and it is not part of the augmented video (dropout phase), (iii) the content of the green clip is replaced with Gaussian noise (dropout phase), and (iv) the frames of the blue clip are replaced with empty frames in the augmented video (dropout phase). The proposed TSD encourages this way, the preservation of some local (clip-level) information and provides certain variations in the global temporal structure of the video, which is beneficial to the generalisation ability of the proposed framework.

\vspace{7pt}
\section{Additional implementation details}
\vspace{8pt}

    All of our models are implemented with the PyTorch~\cite{paszke2019} library. Table~\ref{tab:impl_details} displays all hyperparameters used for the network training and augmentations. 
    
    For global transformations, we use $N_{RAug}=2$ consecutive transformations with $M_{RAug}=9$ magnitude in RandAugment. The frame transformations are applied with probability $p_{overlay}=0.3$ for each text and emoji overlay and with $p_{blur}=0.5$ for blurring. For the temporal transformations, we select TSD with probability $p_{tsd}=0.5$ and the rest transformations with probability $0.1$ each. For TSD, shuffling and dropout are applied with probabilities $p_{shuf}=0.5$ and $p_{drop}=0.3$, respectively, with the latter discarding clips or removing their content with $p_{cont}=0.5$. Finally, the video-in-video transformation is applied with probability $p_{viv}=0.5$, and the donor video is down-sampled with a factor $\lambda_{viv}$ ranging in (0.3, 0.7). 
    
    To enable vectorization during training and facilitate augmentations, we load the original videos with 1 fps, resized to 256 pixels according to their smaller side. Also, to assert that the two augmentation versions overlap, we select a clip of consecutive frames equal to $2T_B$. If the initial video is shorter, it is repeated in the time axis until it reaches the necessary length. We follow the literature and resize/crop the videos to 256/224 pixels.

    All experiments were conducted on a Linux machine with an Intel i9-7900X CPU and two Nvidia 3090 GPUs. Times and storage requirements are the same as the fine-grained attention student reported on the DnS paper~\cite{kordopatis2022}.

    \begin{table}[t]
      \centering
      \scalebox{0.95}{
    \begin{tabular}{|l|c|c|}
        \hline
        \textbf{Parameter} & \textbf{Notation} & \textbf{Value}           \\ \hline
        \multicolumn{3}{|l|}{\textbf{Training process}}                   \\ \hline
        Iterations                  & -               & 30,000            \\ 
        Batch size                  & -               & 64                \\ 
        Optimizer                   & -               & AdamW             \\ 
        Learning rate               & -               & $5\cdot10^{-5}$   \\ 
        Learning rate decay         & -               & cosine            \\ 
        Warmup iterations           & -               & 1,000             \\ 
        Weight decay                & -               & 0.01              \\ \hline
        \# of input frames in batch & $T_B$           & 32                \\ 
        Frame size in batch         & $H_B$,$W_B$     & $224$             \\ \hline
        InfoNCE temperature         & $\tau$          & $0.07$            \\ 
        SSHN loss factor            & $\lambda$       & $3$               \\ 
        Sim. regularization factor  & $r$             & $1$               \\ \hline
        \multicolumn{3}{|l|}{\textbf{Global transformations}}             \\ \hline
        RandAug., \# of transf.     & $N_{RAug}$      & 2                 \\
        RandAug., magnitude         & $M_{RAug}$      & 9                 \\ \hline
        \multicolumn{3}{|l|}{\textbf{Frame transformations}}              \\ \hline
        Overlay prob.               & $p_{overlay}$   & 0.3               \\ 
        Blur prob.                  & $p_{blur}$      & 0.5               \\ \hline
        \multicolumn{3}{|l|}{\textbf{Temporal transformations}}           \\ \hline
        TSD prob.                   & $p_{tsd}$       & 0.5               \\ 
        Fast forward prob.          & $p_{ff}$        & 0.1               \\ 
        Slow motion prob.           & $p_{sm}$        & 0.1               \\ 
        Reverse prob.               & $p_{rev}$       & 0.1               \\ 
        Pause prob.                 & $p_{pau}$       & 0.1               \\ \hline
        TSD shuffle prob.           & $p_{shuf}$      & 0.5               \\ 
        TSD dropout prob.           & $p_{drop}$      & 0.3               \\ 
        TSD content drop prob.      & $p_{cont}$      & 0.5               \\ \hline
        \multicolumn{3}{|l|}{\textbf{Video-in-video}}                     \\ \hline
        ViV prob.                   & $p_{viv}$       & 0.3               \\
        ViV factor range            & $\lambda_{viv}$ & (0.3, 0.7)        \\
        \hline
    \end{tabular}
}
        \caption{Implementation details of the training process and the augmentations parameters.
        \label{tab:impl_details}
      }
    \end{table}

\balance
{
\small
\bibliographystyle{ieee_fullname}
\bibliography{egbib}

\begin{thebibliography}{10}\itemsep=-1pt

\bibitem{agrawal2015learning}
Pulkit Agrawal, Joao Carreira, and Jitendra Malik.
\newblock Learning to see by moving.
\newblock In {\em ICCV}, 2015.

\bibitem{asano2019self}
YM Asano, C Rupprecht, and A Vedaldi.
\newblock Self-labelling via simultaneous clustering and representation
  learning.
\newblock In {\em ICLR}, 2019.

\bibitem{bao2021beit}
Hangbo Bao, Li Dong, Songhao Piao, and Furu Wei.
\newblock Beit: Bert pre-training of image transformers.
\newblock In {\em ICLR}, 2021.

\bibitem{baraldi2018}
Lorenzo Baraldi, Matthijs Douze, Rita Cucchiara, and Herv{\'e} J{\'e}gou.
\newblock {LAMV}: Learning to align and match videos with kernelized temporal
  layers.
\newblock In {\em CVPR}, 2018.

\bibitem{bishay2019}
Mina Bishay, Georgios Zoumpourlis, and I. Patras.
\newblock {TARN}: Temporal attentive relation network for few-shot and
  zero-shot action recognition.
\newblock In {\em BMVC}, 2019.

\bibitem{cai2011}
Yang Cai, Linjun Yang, Wei Ping, Fei Wang, Tao Mei, Xian-Sheng Hua, and Shipeng
  Li.
\newblock Million-scale near-duplicate video retrieval system.
\newblock In {\em ACM MM}, 2011.

\bibitem{caron2018deep}
Mathilde Caron, Piotr Bojanowski, Armand Joulin, and Matthijs Douze.
\newblock Deep clustering for unsupervised learning of visual features.
\newblock In {\em ECCV}, 2018.

\bibitem{caron2020unsupervised}
Mathilde Caron, Ishan Misra, Julien Mairal, Priya Goyal, Piotr Bojanowski, and
  Armand Joulin.
\newblock Unsupervised learning of visual features by contrasting cluster
  assignments.
\newblock {\em NeurIPS}, 2020.

\bibitem{caron2021emerging}
Mathilde Caron, Hugo Touvron, Ishan Misra, Herv{\'e} J{\'e}gou, Julien Mairal,
  Piotr Bojanowski, and Armand Joulin.
\newblock Emerging properties in self-supervised vision transformers.
\newblock In {\em ICCV}, 2021.

\bibitem{chen2020simple}
Ting Chen, Simon Kornblith, Mohammad Norouzi, and Geoffrey Hinton.
\newblock A simple framework for contrastive learning of visual
  representations.
\newblock In {\em ICML}, 2020.

\bibitem{chen2021exploring}
Xinlei Chen and Kaiming He.
\newblock Exploring simple siamese representation learning.
\newblock In {\em CVPR}, 2021.

\bibitem{chou2015}
Chien-Li Chou, Hua-Tsung Chen, and Suh-Yin Lee.
\newblock Pattern-based near-duplicate video retrieval and localization on
  web-scale videos.
\newblock {\em IEEE TMM}, 2015.

\bibitem{chum2007}
Ondrej Chum, James Philbin, Josef Sivic, Michael Isard, and Andrew Zisserman.
\newblock Total recall: Automatic query expansion with a generative feature
  model for object retrieval.
\newblock In {\em ICCV}, 2007.

\bibitem{cubuk2020}
Ekin~D Cubuk, Barret Zoph, Jonathon Shlens, and Quoc~V Le.
\newblock {RandAugment}: Practical automated data augmentation with a reduced
  search space.
\newblock In {\em CVPRW}, 2020.

\bibitem{deng2009imagenet}
Jia Deng, Wei Dong, Richard Socher, Li-Jia Li, Kai Li, and Li Fei-Fei.
\newblock Imagenet: A large-scale hierarchical image database.
\newblock In {\em CVPR}, 2009.

\bibitem{doersch2015unsupervised}
Carl Doersch, Abhinav Gupta, and Alexei~A Efros.
\newblock Unsupervised visual representation learning by context prediction.
\newblock In {\em ICCV}, 2015.

\bibitem{doersch2017multi}
Carl Doersch and Andrew Zisserman.
\newblock Multi-task self-supervised visual learning.
\newblock In {\em ICCV}, 2017.

\bibitem{douze2010}
Matthijs Douze, Herv{\'e} J{\'e}gou, and Cordelia Schmid.
\newblock An image-based approach to video copy detection with spatio-temporal
  post-filtering.
\newblock {\em IEEE TMM}, 2010.

\bibitem{douze2021}
Matthijs Douze, Giorgos Tolias, Ed Pizzi, Zo{\"e} Papakipos, Lowik Chanussot,
  Filip Radenovic, Tomas Jenicek, Maxim Maximov, Laura Leal-Taix{\'e}, Ismail
  Elezi, et~al.
\newblock The 2021 image similarity dataset and challenge.
\newblock In {\em arXiv:2106.09672}, 2021.

\bibitem{feng2018}
Yang Feng, Lin Ma, Wei Liu, Tong Zhang, and Jiebo Luo.
\newblock Video re-localization.
\newblock In {\em ECCV}, 2018.

\bibitem{fernando2017self}
Basura Fernando, Hakan Bilen, Efstratios Gavves, and Stephen Gould.
\newblock Self-supervised video representation learning with odd-one-out
  networks.
\newblock In {\em CVPR}, 2017.

\bibitem{gao2017}
Zhanning Gao, Gang Hua, Dongqing Zhang, Nebojsa Jojic, Le Wang, Jianru Xue, and
  Nanning Zheng.
\newblock {ER3}: A unified framework for event retrieval, recognition and
  recounting.
\newblock In {\em CVPR}, 2017.

\bibitem{grill2020bootstrap}
Jean-Bastien Grill, Florian Strub, Florent Altch{\'e}, Corentin Tallec, Pierre
  Richemond, Elena Buchatskaya, Carl Doersch, Bernardo Avila~Pires, Zhaohan
  Guo, Mohammad Gheshlaghi~Azar, et~al.
\newblock Bootstrap your own latent-a new approach to self-supervised learning.
\newblock {\em NeurIPS}, 2020.

\bibitem{han2020self}
Tengda Han, Weidi Xie, and Andrew Zisserman.
\newblock Self-supervised co-training for video representation learning.
\newblock {\em NeurIPS}, 2020.

\bibitem{han2021}
Zhen Han, Xiangteng He, Mingqian Tang, and Yiliang Lv.
\newblock Video similarity and alignment learning on partial video copy
  detection.
\newblock In {\em ACM MM}, 2021.

\bibitem{he2022masked}
Kaiming He, Xinlei Chen, Saining Xie, Yanghao Li, Piotr Doll{\'a}r, and Ross
  Girshick.
\newblock Masked autoencoders are scalable vision learners.
\newblock In {\em CVPR}, 2022.

\bibitem{he2020momentum}
Kaiming He, Haoqi Fan, Yuxin Wu, Saining Xie, and Ross Girshick.
\newblock Momentum contrast for unsupervised visual representation learning.
\newblock In {\em CVPR}, 2020.

\bibitem{he2016}
Kaiming He, Xiangyu Zhang, Shaoqing Ren, and Jian Sun.
\newblock Deep residual learning for image recognition.
\newblock In {\em CVPR}, 2016.

\bibitem{he2022vcsl}
Sifeng He, Xudong Yang, Chen Jiang, Gang Liang, Wei Zhang, Tan Pan, Qing Wang,
  Furong Xu, Chunguang Li, JinXiong Liu, et~al.
\newblock A large-scale comprehensive dataset and copy-overlap aware evaluation
  protocol for segment-level video copy detection.
\newblock In {\em CVPR}, 2022.

\bibitem{he2023transvcl}
Sifeng He, He Yue, Minlong Lu, et~al.
\newblock {TransVCL}: Attention-enhanced video copy localization network with
  flexible supervision.
\newblock In {\em AAAI}, 2023.

\bibitem{he2022}
Xiangteng He, Yulin Pan, Mingqian Tang, Yiliang Lv, and Yuxin Peng.
\newblock Learn from unlabeled videos for near-duplicate video retrieval.
\newblock In {\em ACM SIGIR}, 2022.

\bibitem{henaff2020data}
Olivier Henaff.
\newblock Data-efficient image recognition with contrastive predictive coding.
\newblock In {\em ICML}, 2020.

\bibitem{hochreiter1997}
Sepp Hochreiter and J{\"u}rgen Schmidhuber.
\newblock Long short-term memory.
\newblock {\em Neural computation}, 1997.

\bibitem{huang1999}
Jing Huang, S Ravi~Kumar, Mandar Mitra, Wei-Jing Zhu, and Ramin Zabih.
\newblock Spatial color indexing and applications.
\newblock {\em IJCV}, 1999.

\bibitem{huang2010}
Zi Huang, Heng~Tao Shen, Jie Shao, Bin Cui, and Xiaofang Zhou.
\newblock Practical online near-duplicate subsequence detection for continuous
  video streams.
\newblock {\em IEEE TMM}, 2010.

\bibitem{jayaraman2015learning}
Dinesh Jayaraman and Kristen Grauman.
\newblock Learning image representations tied to ego-motion.
\newblock In {\em ICCV}, 2015.

\bibitem{jiang2021}
Chen Jiang, Kaiming Huang, Sifeng He, Xudong Yang, Wei Zhang, Xiaobo Zhang,
  Yuan Cheng, Lei Yang, Qing Wang, Furong Xu, et~al.
\newblock Learning segment similarity and alignment in large-scale content
  based video retrieval.
\newblock In {\em ACM MM}, 2021.

\bibitem{jiang2014}
Yu-Gang Jiang, Yudong Jiang, and Jiajun Wang.
\newblock {VCDB}: A large-scale database for partial copy detection in videos.
\newblock In {\em ECCV}, 2014.

\bibitem{jiang2016}
Yu-Gang Jiang and Jiajun Wang.
\newblock Partial copy detection in videos: A benchmark and an evaluation of
  popular methods.
\newblock {\em IEEE TBD}, 2016.

\bibitem{kim2019self}
Dahun Kim, Donghyeon Cho, and In~So Kweon.
\newblock Self-supervised video representation learning with space-time cubic
  puzzles.
\newblock In {\em AAAI}, 2019.

\bibitem{kordopatis2017a}
Giorgos Kordopatis-Zilos, Symeon Papadopoulos, Ioannis Patras, and Ioannis
  Kompatsiaris.
\newblock Near-duplicate video retrieval by aggregating intermediate cnn
  layers.
\newblock In {\em MMM}, 2017.

\bibitem{kordopatis2017b}
Giorgos Kordopatis-Zilos, Symeon Papadopoulos, Ioannis Patras, and Ioannis
  Kompatsiaris.
\newblock Near-duplicate video retrieval with deep metric learning.
\newblock In {\em ICCVW}, 2017.

\bibitem{kordopatis2019a}
Giorgos Kordopatis-Zilos, Symeon Papadopoulos, Ioannis Patras, and Ioannis
  Kompatsiaris.
\newblock {FIVR: Fine-grained Incident Video Retrieval}.
\newblock {\em IEEE TMM}, 2019.

\bibitem{kordopatis2019b}
Giorgos Kordopatis-Zilos, Symeon Papadopoulos, Ioannis Patras, and Ioannis
  Kompatsiaris.
\newblock {ViSiL}: Fine-grained spatio-temporal video similarity learning.
\newblock In {\em ICCV}, 2019.

\bibitem{kordopatis2022}
Giorgos Kordopatis-Zilos, Christos Tzelepis, Symeon Papadopoulos, Ioannis
  Kompatsiaris, and Ioannis Patras.
\newblock {DnS: Distill-and-Select for Efficient and Accurate Video Indexing
  and Retrieval}.
\newblock {\em IJCV}, 2022.

\bibitem{law2007video}
Julien Law-To, Li Chen, Alexis Joly, Ivan Laptev, Olivier Buisson, Valerie
  Gouet-Brunet, Nozha Boujemaa, and Fred Stentiford.
\newblock Video copy detection: a comparative study.
\newblock In {\em ACM CIVR}, 2007.

\bibitem{lee2020}
Hyodong Lee, Joonseok Lee, Joe Yue-Hei Ng, and Paul Natsev.
\newblock Large scale video representation learning via relational graph
  clustering.
\newblock In {\em CVPR}, 2020.

\bibitem{lee2018}
Joonseok Lee, Sami Abu-El-Haija, Balakrishnan Varadarajan, and Apostol Natsev.
\newblock Collaborative deep metric learning for video understanding.
\newblock In {\em ACM SIGKDD}, 2018.

\bibitem{li2022dual}
Pandeng Li, Hongtao Xie, Jiannan Ge, Lei Zhang, Shaobo Min, and Yongdong Zhang.
\newblock Dual-stream knowledge-preserving hashing for unsupervised video
  retrieval.
\newblock In {\em ECCV}, 2022.

\bibitem{li2021}
Shuyan Li, Xiu Li, Jiwen Lu, and Jie Zhou.
\newblock Self-supervised video hashing via bidirectional transformers.
\newblock In {\em CVPR}, 2021.

\bibitem{loshchilov2016sgdr}
Ilya Loshchilov and Frank Hutter.
\newblock {SGDR}: Stochastic gradient descent with warm restarts.
\newblock {\em ICLR}, 2016.

\bibitem{loshchilov2018decoupled}
Ilya Loshchilov and Frank Hutter.
\newblock Decoupled weight decay regularization.
\newblock In {\em ICLR}, 2018.

\bibitem{misra2016shuffle}
Ishan Misra, C~Lawrence Zitnick, and Martial Hebert.
\newblock Shuffle and learn: unsupervised learning using temporal order
  verification.
\newblock In {\em ECCV}, 2016.

\bibitem{ng2022vrag}
Kennard Ng, Ser-Nam Lim, and Gim~Hee Lee.
\newblock {VRAG}: Region attention graphs for content-based video retrieval.
\newblock In {\em arXiv:2205.09068}, 2022.

\bibitem{noroozi2016unsupervised}
Mehdi Noroozi and Paolo Favaro.
\newblock Unsupervised learning of visual representations by solving jigsaw
  puzzles.
\newblock In {\em ECCV}, 2016.

\bibitem{ojala1996}
Timo Ojala, Matti Pietik{\"a}inen, and David Harwood.
\newblock A comparative study of texture measures with classification based on
  featured distributions.
\newblock {\em PR}, 1996.

\bibitem{oord2018representation}
Aaron van~den Oord, Yazhe Li, and Oriol Vinyals.
\newblock Representation learning with contrastive predictive coding.
\newblock In {\em arXiv:1807.03748}, 2018.

\bibitem{paszke2019}
Adam Paszke, Sam Gross, Francisco Massa, Adam Lerer, James Bradbury, Gregory
  Chanan, Trevor Killeen, Zeming Lin, Natalia Gimelshein, Luca Antiga, et~al.
\newblock {PyTorch}: An imperative style, high-performance deep learning
  library.
\newblock In {\em NeurIPS}, 2019.

\bibitem{pathak2016context}
Deepak Pathak, Philipp Krahenbuhl, Jeff Donahue, Trevor Darrell, and Alexei~A
  Efros.
\newblock Context encoders: Feature learning by inpainting.
\newblock In {\em CVPR}, 2016.

\bibitem{pizzi2022}
Ed Pizzi, Sreya~Dutta Roy, Sugosh~Nagavara Ravindra, Priya Goyal, and Matthijs
  Douze.
\newblock A self-supervised descriptor for image copy detection.
\newblock In {\em CVPR}, 2022.

\bibitem{poullot2015}
S{\'e}bastien Poullot, Shunsuke Tsukatani, Anh Phuong~Nguyen, Herv{\'e}
  J{\'e}gou, and Shin'Ichi Satoh.
\newblock Temporal matching kernel with explicit feature maps.
\newblock In {\em ACM MM}, 2015.

\bibitem{qian2021spatiotemporal}
Rui Qian, Tianjian Meng, Boqing Gong, Ming-Hsuan Yang, Huisheng Wang, Serge
  Belongie, and Yin Cui.
\newblock Spatiotemporal contrastive video representation learning.
\newblock In {\em CVPR}, 2021.

\bibitem{recasens2021broaden}
Adria Recasens, Pauline Luc, Jean-Baptiste Alayrac, Luyu Wang, Florian Strub,
  Corentin Tallec, Mateusz Malinowski, Viorica P{\u{a}}tr{\u{a}}ucean, Florent
  Altch{\'e}, Michal Valko, et~al.
\newblock Broaden your views for self-supervised video learning.
\newblock In {\em ICCV}, 2021.

\bibitem{revaud2013}
J{\'e}r{\^o}me Revaud, Matthijs Douze, Cordelia Schmid, and Herv{\'e}
  J{\'e}gou.
\newblock Event retrieval in large video collections with circulant temporal
  encoding.
\newblock In {\em CVPR}, 2013.

\bibitem{sablayrolles2018}
Alexandre Sablayrolles, Matthijs Douze, Cordelia Schmid, and Herv{\'e}
  J{\'e}gou.
\newblock Spreading vectors for similarity search.
\newblock In {\em ICLR}, 2018.

\bibitem{shang2010}
Lifeng Shang, Linjun Yang, Fei Wang, Kwok-Ping Chan, and Xian-Sheng Hua.
\newblock Real-time large scale near-duplicate web video retrieval.
\newblock In {\em ACM MM}, 2010.

\bibitem{shao2021}
Jie Shao, Xin Wen, Bingchen Zhao, and Xiangyang Xue.
\newblock Temporal context aggregation for video retrieval with contrastive
  learning.
\newblock In {\em WACV}, 2021.

\bibitem{sivic2003}
Josef Sivic and Andrew Zisserman.
\newblock {Video Google}: A text retrieval approach to object matching in
  videos.
\newblock In {\em CVPR}, 2003.

\bibitem{song2011}
Jingkuan Song, Yi Yang, Zi Huang, Heng~Tao Shen, and Richang Hong.
\newblock Multiple feature hashing for real-time large scale near-duplicate
  video retrieval.
\newblock In {\em ACM MM}, 2011.

\bibitem{song2018}
Jingkuan Song, Hanwang Zhang, Xiangpeng Li, Lianli Gao, Meng Wang, and Richang
  Hong.
\newblock Self-supervised video hashing with hierarchical binary auto-encoder.
\newblock {\em IEEE TIP}, 2018.

\bibitem{tan2009}
Hung-Khoon Tan, Chong-Wah Ngo, Richard Hong, and Tat-Seng Chua.
\newblock Scalable detection of partial near-duplicate videos by
  visual-temporal consistency.
\newblock In {\em ACM MM}, 2009.

\bibitem{tian2020contrastive}
Yonglong Tian, Dilip Krishnan, and Phillip Isola.
\newblock Contrastive multiview coding.
\newblock In {\em ECCV}, 2020.

\bibitem{tolias2016}
Giorgos Tolias, Ronan Sicre, and Herv{\'e} J{\'e}gou.
\newblock Particular object retrieval with integral max-pooling of cnn
  activations.
\newblock In {\em ICLR}, 2016.

\bibitem{tong2022videomae}
Zhan Tong, Yibing Song, Jue Wang, and Limin Wang.
\newblock Videomae: Masked autoencoders are data-efficient learners for
  self-supervised video pre-training.
\newblock In {\em NeurIPS}, 2022.

\bibitem{vaswani2017}
Ashish Vaswani, Noam Shazeer, Niki Parmar, Jakob Uszkoreit, Llion Jones,
  Aidan~N Gomez, Lukasz Kaiser, and Illia Polosukhin.
\newblock Attention is all you need.
\newblock In {\em NeurIPS}, 2017.

\bibitem{wang2021}
Kuan-Hsun Wang, Chia-Chun Cheng, Yi-Ling Chen, Yale Song, and Shang-Hong Lai.
\newblock Attention-based deep metric learning for near-duplicate video
  retrieval.
\newblock In {\em ICPR}, 2021.

\bibitem{wang2017}
Ling Wang, Yu Bao, Haojie Li, Xin Fan, and Zhongxuan Luo.
\newblock Compact cnn based video representation for efficient video copy
  detection.
\newblock In {\em MMM}, 2017.

\bibitem{wang2022bevt}
Rui Wang, Dongdong Chen, Zuxuan Wu, Yinpeng Chen, Xiyang Dai, Mengchen Liu,
  Yu-Gang Jiang, Luowei Zhou, and Lu Yuan.
\newblock Bevt: Bert pretraining of video transformers.
\newblock In {\em CVPR}, 2022.

\bibitem{wu2007}
Xiao Wu, Alexander~G Hauptmann, and Chong-Wah Ngo.
\newblock Practical elimination of near-duplicates from web video search.
\newblock In {\em ACM MM}, 2007.

\bibitem{yang2016}
Zichao Yang, Diyi Yang, Chris Dyer, Xiaodong He, Alex Smola, and Eduard Hovy.
\newblock Hierarchical attention networks for document classification.
\newblock In {\em nAC-ACL}, 2016.

\bibitem{yuan2020}
Li Yuan, Tao Wang, Xiaopeng Zhang, Francis~EH Tay, Zequn Jie, Wei Liu, and
  Jiashi Feng.
\newblock Central similarity quantization for efficient image and video
  retrieval.
\newblock In {\em CVPR}, 2020.

\bibitem{yun2019}
Sangdoo Yun, Dongyoon Han, Seong~Joon Oh, Sanghyuk Chun, Junsuk Choe, and
  Youngjoon Yoo.
\newblock Cutmix: Regularization strategy to train strong classifiers with
  localizable features.
\newblock In {\em ICCV}, 2019.

\bibitem{zhang2016colorful}
Richard Zhang, Phillip Isola, and Alexei~A Efros.
\newblock Colorful image colorization.
\newblock In {\em ECCV}, 2016.

\end{thebibliography}
}

\end{document}